\documentclass[10pt,twocolumn,letterpaper]{article}

\usepackage{cvpr}              

\usepackage{graphicx}
\usepackage{amsmath}
\usepackage[makeroom]{cancel}
\usepackage{amssymb}
\usepackage{booktabs}
\usepackage[accsupp]{axessibility}

\usepackage[pagebackref,breaklinks,colorlinks]{hyperref}

\usepackage[capitalize]{cleveref}
\crefname{section}{Sec.}{Secs.}
\Crefname{section}{Section}{Sections}
\Crefname{table}{Table}{Tables}
\crefname{table}{Tab.}{Tabs.}

\def\cvprPaperID{0} 
\def\confName{CVPR}
\def\confYear{2022}

\begin{document}

\title{PIE-Net: Photometric Invariant Edge Guided Network for Intrinsic Image Decomposition}

\author{Partha Das\\
University of Amsterdam,\\
3DUniversum\\
Amsterdam, The Netherlands\\
{\tt\small p.das@uva.nl}
\and
Sezer Karaoglu\\
University of Amsterdam,\\
3DUniversum\\
Amsterdam, The Netherlands\\
{\tt\small s.karaoglu@3duniversum.com}
\and
Theo Gevers\\
University of Amsterdam,\\
3DUniversum\\
Amsterdam, The Netherlands\\
{\tt\small Th.Gevers@uva.nl}
}
\maketitle

\begin{abstract}
   Intrinsic image decomposition is the process of recovering the image formation components (reflectance and shading) from an image. Previous methods employ either explicit priors to constrain the problem or implicit constraints as formulated by their losses (deep learning). These methods can be negatively influenced by strong illumination conditions causing shading-reflectance leakages.
   
   Therefore, in this paper, an end-to-end edge-driven hybrid CNN approach is proposed for intrinsic image decomposition. Edges correspond to illumination invariant gradients. To handle hard negative illumination transitions, a hierarchical approach is taken including global and local refinement layers. We make use of attention layers to further strengthen the learning process. 
   
   An extensive ablation study and large scale experiments are conducted showing that it is beneficial for edge-driven hybrid IID networks to make use of illumination invariant descriptors and that separating global and local cues helps in improving the performance of the network. Finally, it is shown that the proposed method obtains state of the art performance and is able to generalise well to real world images. The project page with pretrained models,  finetuned models and network code can be found at https://ivi.fnwi.uva.nl/cv/pienet/.
\end{abstract}

\section{Introduction}

Intrinsic Image Decomposition (IID) is the process of recovering the image formation components such as reflectance (albedo) and shading (illumination) from an image. The reflectance image can be used for albedo texture edits~\cite{Ye2014,Meka2016,Beigpour2011ObjectRB}, fabric recolouring~\cite{Xu2019} or semantic segmentation~\cite{Baslamisli2018ECCV}. As for shading, the illumination image can be used for relighting~\cite{shu2017} or shape-from-shading i.e. estimating the shape/geometry of objects or scenes~\cite{Wada1995,henderson2019}.

\begin{figure*}
    \centering
    \includegraphics[width=\linewidth,height=0.4\linewidth]{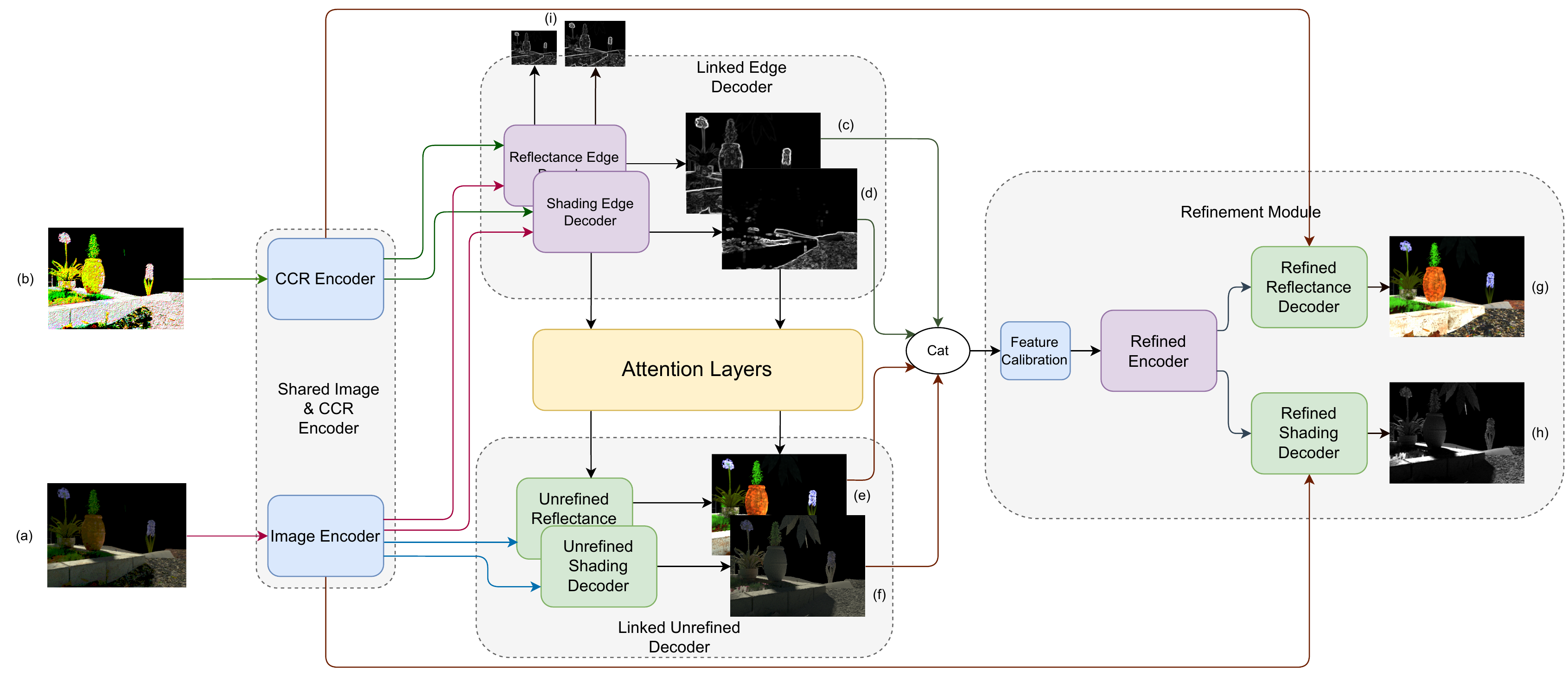}
    \caption{Overview of the proposed Network. The architecture consists of 4 sub-modules denoted by dotted boxes. Inputs to the network are (a) a $RGB$ image and (b) a CCR image. The CCR image is computed from (a). The outputs of the networks are: (c) the reflectance edge, (d) the shading edge, (e) the unrefined reflectance prediction, (f) the unrefined shading prediction, (g) the final refined reflectance, (h) the final refined shading, and (i) the scaled edge outputs @(64, 128).}
    \label{fig:net_over_view}
\end{figure*}

The problem of IID is inherently ill-defined. Therefore, previous IID approaches employ priors to constrain the problem. Retinex ~\cite{Land1971} is based on gradient information derived from images where shading variations correspond to small (soft) gradients and reflectance transitions to larger (stronger) ones. Other constraints are explored by~\cite{Barron2015}, like piece-wise constancy, parsimony of reflectance, shading smoothness, etc. Other approaches include global sparsity priors on the palette of colours (albedo's) and modelling the problem as latent variable Random Fields by~\cite{Gehler2011}. However, these explicitly imposed constraints (i.e. assumptions about the world) may limit the applicability of these methods. Recently, deep learning based methods are proposed ~\cite{Narihia2015,Shi2017}. These methods are based on implicit constraints as formulated by losses, and multiple datasets or image sequences~\cite{Li2018ECCV,Weiss2001}. However, these approaches are purely data-driven and therefore they may be limited in their generalisation abilities (dataset bias). Traditional constraints and deep learning approaches are combined by~\cite{Fan2018} by means of image edge guidance. However, edge-driven hybrid methods can be influenced by strong illumination conditions. For example, in case of strong shadows, the network may classify shadows as being reflectance edges. This happens when the gradient assumption is violated: illumination changes correspond to soft gradients and reflectance transitions to hard ones. This leads to the well-known problem of shading-reflectance leakage i.e. illumination (strong shadow/shading) transitions which are interpreted/classified as albedo transitions also called hard negative illumination transitions, or simply hard (illumination) negatives.

Therefore, in this paper, an edge-driven hybrid CNN approach is proposed using gradients based on illumination invariant descriptors i.e. Cross Color Ratios (CCR)~\cite{Gevers1999}. CCR are illumination (including shadows and shading patterns) invariant gradients and hence only dependent on albedo changes. To solve for hard negative illumination transitions, a hierarchical CNN is proposed including global and local refinement layers. The global layer ensures a smooth image decomposition eliminating (soft) negative illumination transitions and therefore minimising shading-reflectance misclassification. The local refinement will eliminate the hard (illumination) negatives. A two-staged approach is beneficial because by adding an incremental parameter space that is conditioned on the previous step, the network is no longer required to model both strong and soft illumination patterns in a single process, but rather a refinement elimination process to remove the hard negatives. The proposed method implicitly encodes the image intrinsics within the network without the need of manual thresholding. Since intrinsic components are spatially dependent, spatial attention layers are included. This allows the network to focus on image areas containing hard negatives. Fig.~\ref{fig:net_over_view} provides an overview of the proposed network. 

In summary, our contributions are as follows:
\begin{itemize}
    \item An end-to-end edge-driven hybrid approach is proposed for intrinsic image decomposition using gradients based on illumination invariant descriptors.
    \item To solve for hard negative illumination transitions, a hierarchical approach is taken including global and local refinement layers.
    \item It is shown that separating the parameter space in a global and local space, rather than a unified parameter space, outperforms single parameter space learning.
    \item It is shown that the proposed algorithm is able to achieve state of the art performance and is able to generalise well to real world images.
\end{itemize}

\section{Related Work}

Earlier methods on intrinsic image decomposition are mainly focused on exploring hand-crafted priors to reduce the solution space.~\cite{Land1971} argues that sharp gradient changes belong to reflectance changes, while soft transitions correspond to illumination patterns. ~\cite{Gevers1999} proposes the Cross Color Ratios (CCR). CCR are illumination (including shadows and shading patterns) invariant gradients and hence only dependent on albedo transitions. Surprisingly, they have not been employed in the domain of IID. 

Other methods, like~\cite{Shen2008} explore texture cues where image areas having the same reflectance values should also have the same intensity values.~\cite{Gehler2011, Barron2015} adds multiple constraints like piece-wise consistency for reflectance and smoothness priors for shading. Other constraints include textures~\cite{Shen2011,Gehler2011}, depth cues~\cite{Barron2013,Lee2012}, infrared priors~\cite{Cheng2019} and surface normals~\cite{Jeon2014}. Shading is also decomposed into geometry and illumination by regularising them individually~\cite{Chen2013}. Different optimisation frameworks are explored by~\cite{Bell2014}, where Conditional Random Fields are used to optimise pairwise reflectances. More implicit reflectance constraints like multiple frames are explored by~\cite{Weiss2001, Matsushita2004}. User annotated priors are explored by~\cite{Bousseau2009, Bonneel2014, Bell2014, Narihira2015-2}. For these methods, the application domains are often limited to single objects. 

\cite{Narihia2015} is the first to represent the problem of IID using a CNN.~\cite{Shi2017} expands this by introducing skip connections and inter-component connections to enforce component  inter-dependence.~\cite{Janner2017,Zhou2019,Baslamisli2019} explore decomposing the shading component further into shadows, direct and ambient lighting cues. Furthermore, ~\cite{Baslamisli2018CVPR} introduces RetiNet by parameterising the Retinex algorithm as an end-to-end learnable framework. Other approaches include Laplacian pyramids based on scale space learning~\cite{Cheng2018}, adversarial residual networks~\cite{Lettry2018}, inverse rendering~\cite{Sengupta2019} and image edge guidance~\cite{Fan2018}. Finally,~\cite{Li2018ECCV} combines multiple constraints as loss functions and trains an end-to-end system using 4 different datasets simultaneously.~\cite{Li2020} models the problem as differentiable rendering layers and trains it in an end-to-end manner with supervision on reflectance, roughness, normals, depths, etc. Apart from the supervised methods, unsupervised learning approaches for IID are also explored in~\cite{Liu2020, Xu2020, Li2018ECCV, Yu2019}. Although the results of these approaches are promising, these methods are not always able to fully disentangle (strong) shading and reflectance transitions i.e shading-reflectance leakage problem.

\section{Methodology}

\subsection{Illumination Invariant Gradients}
Given an $RGB$ image and two neighbouring pixels $p_1$ \& $p_2$. Then, the Cross Color Ratios are defined by:

\begin{equation} \label{eq:cross_colour_ratios}
\begin{aligned}
        M_{RG} = \;\frac{R_{p_1}\;G_{p_2}}{R_{p_2}\;G_{p_1}}\;,
        M_{RB} = \;\frac{R_{p_1}\;B_{p_2}}{R_{p_2}\;B_{p_1}}\;,
        M_{GB} = \;\frac{G_{p_1}\;B_{p_2}}{G_{p_2}\;B_{p_1}}\;,
\end{aligned}
\end{equation}

\noindent where $M_{RG}, M_{RB}, M_{GB}$ are the CCR for the ($R,~G$), ($R,~B$) \& ($G,~B$) channel pairs, respectively. Taking the logarithm on both sides of the equation, we obtain:

\begin{equation} \label{eq:cross_colour_ratios_2}
\begin{aligned}
        log(M_{RG}) &= \;log(R_{p_1}\;G_{p_2})\; - \;log(R_{p_2}\;G_{p_1})\;,\\
        log(M_{RB}) &= \;log(R_{p_1}\;B_{p_2})\; - \;log(R_{x_2}\;B_{x_1})\;,\\
        log(M_{GB}) &= \;log(G_{p_1}\;B_{p_2})\; - \;log(G_{p_2}\;B_{p_1})\;.
\end{aligned}
\end{equation}

\noindent Let the image formation process be modelled by~\cite{Shafer1985}:

\begin{equation} \label{eq:imf}
\begin{split}
I = m(\vec{n}, \vec{l}) \int_{\omega}^{}\;e(\lambda)\;
    \rho_{b}(\lambda)\; f(\lambda)\; \mathrm{d}\lambda \;,
\end{split}
\end{equation}

\noindent where, $I$ is the captured image; $\lambda$ is the incoming light wavelength within the visible spectrum $\omega$; $m$ is a function depending on the object geometry and light sources; $\vec{n}$ denotes the surface normal and $\vec{l}$ corresponds to the light source direction. $f$ indicates the spectral camera sensitivity and $e$ describes the spectral power distribution of the light source. Reflectance is denoted by $\rho$ and is related to the albedo/colour of the object. Discretising the model, we obtain:

\begin{equation}
    \begin{aligned}
        C_{p_1} = m(\vec{n},\;\vec{l})\;e^{C_{p_1}}(\lambda)\;
        \rho^{C_{p_1}}(\lambda)\;,
    \end{aligned}{}\label{eq:chan_eq}
\end{equation}{}

\noindent where $C_{p_1}$ is colour channel $C$ for pixel $p_1$ for a $RGB$ image.

For two neighbouring pixels $p_1$ and $p_2$, the same illumination conditions can be assumed since they are very close to each other. Hence:

\begin{equation}
    e^{C_{p_1}} = e^{C_{p_2}}\;,
\end{equation}

Combining Eq.~\eqref{eq:cross_colour_ratios_2} and Eq~\eqref{eq:chan_eq} results in:

\begin{equation} \label{eq:M_sanity}
\begin{aligned}
        log(M_{RG}) &=\;\;\;\;log(R_{p_1}\;G_{p_2})\;-\;log(R_{p_2}\;G_{p_1})\;,\\
        log(M_{RG}) &=\;\;\;\;log(R_{p_1}) \;+\; log(G_{p_2})\;\\
          &\;\;\;-\;log(R_{p_2})\;-\; log(G_{p_1})\;,\\
        log(M_{RG}) &=\;\;\;\;log(\rho^{R_{p_1}}(\lambda))\;+\;log(\rho^{G_{p_2}}(\lambda)) \\
          &\;\;\;-\;log(\rho^{R_{p_2}}(\lambda))\;-\;log(\rho^{G_{p_1}}(\lambda))\;.\\
\end{aligned}
\end{equation}

\noindent Hence, CCR are illumination invariant differences and they are only dependent on the reflectance transitions. To reconstruct the intrinsic (shading and reflectance) images from these edges, a CNN is proposed.

\begin{figure}[ht]
\begin{center}
\includegraphics[width=0.9\linewidth,height=0.5\linewidth]{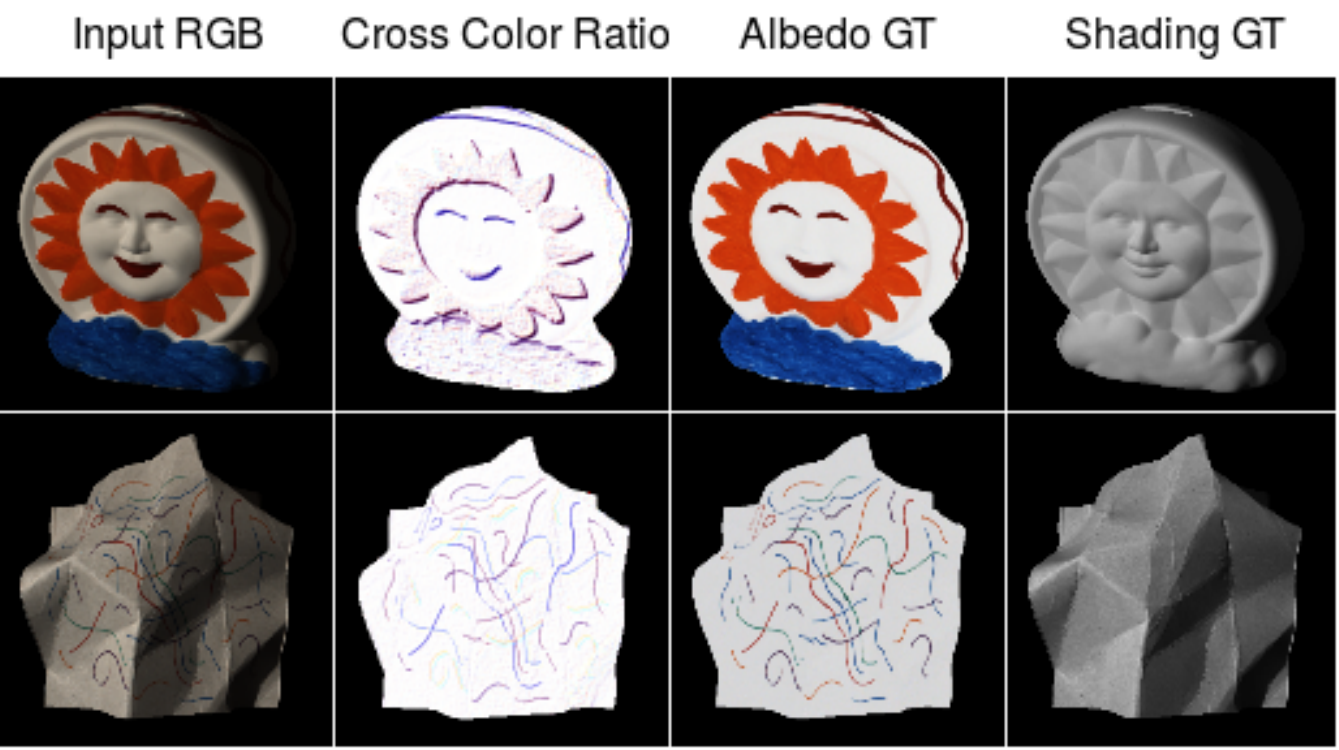}
\end{center}
   \caption{The CCR value becomes 1 where the reflectance is constant. CCR changes correspond to reflectance changes. CCR are illuminantion invariant. Images are gamma corrected for visualisation purposes.}
\label{fig:ccr}
\end{figure}

Fig.~\ref{fig:ccr} shows that CCR, computed for two images, correspond to reflectance changes. They are not dependent on illumination changes. However, noisy regions and strong illumination may introduce shading-reflectance leakage transitions. Therefore, we propose a hierarchical CNN including global and local refinement layers. The refinement layers are used to cope with hard negative illumination transitions. To this end, CCR are integrated in the network to steer the learning process through the use of encoded features in: 1) global refinement through edge prediction, and 2) local patch-wise consistent refinement.

\subsection{Network Architecture \& Details}
\label{sec:netarch}

The network architecture is composed of four sub-components: 1) A Shared Image \& CCR Encoder, 2) Linked Edge Decoder, 3) Unrefined Decoder and 4) Local Refinement Module. The entire network is trained end-to-end. 

\begin{figure}
    \centering
    \includegraphics[width=0.8\linewidth]{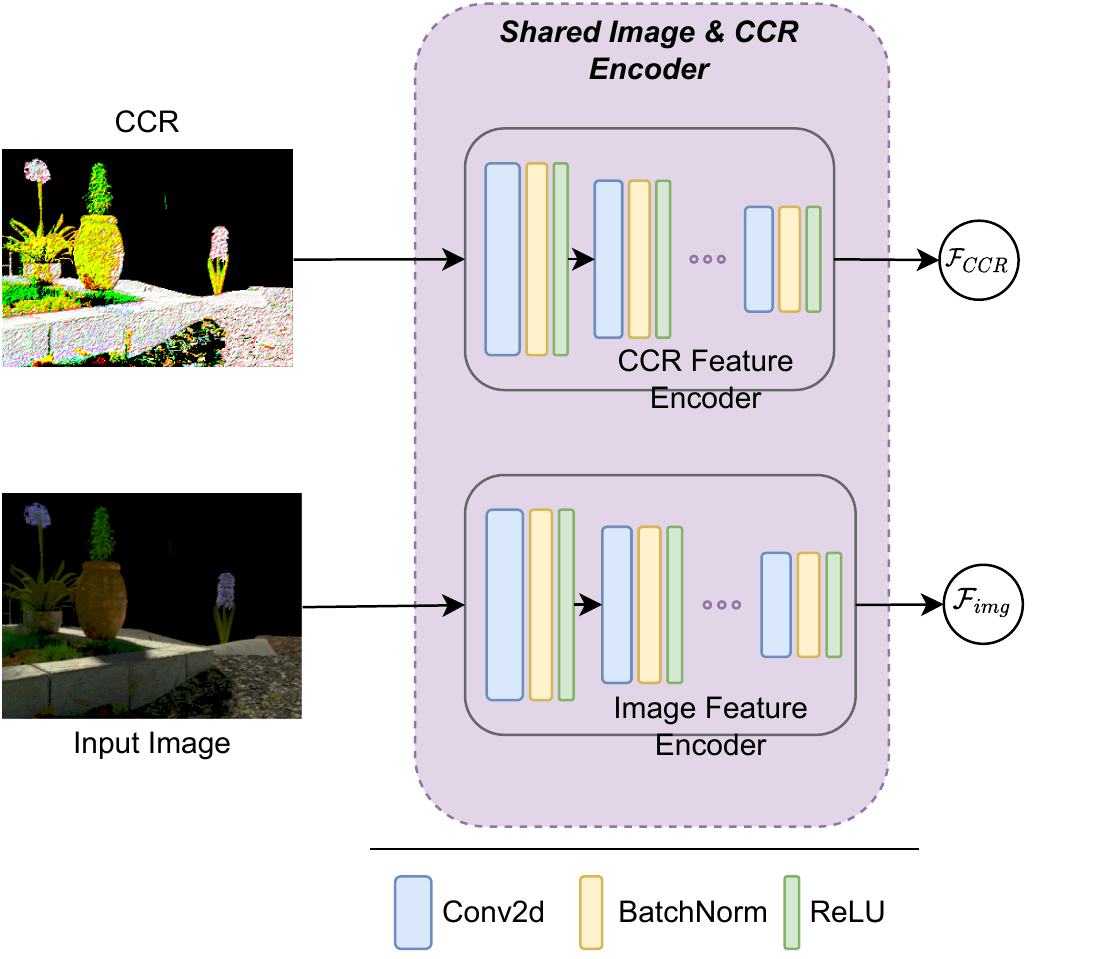}
    \caption{Overview of the shared encoder. The input $RGB$ images and the corresponding CCR are generated by independent encoders. These encoded features are then passed on to the rest of the network as a guidance for the global and local layers.}
    \label{fig:shared_enc}
\end{figure}

\paragraph{Shared Image \& CCR Encoder:}Fig.~\ref{fig:shared_enc} shows the proposed image and CCR encoders. The input and corresponding CCR image are encoded through two separate encoders. This allows the CCR Encoder to learn illumination invariant reflectance transition feature ($\mathcal{F}_{CCR}$), while the image encoder learns an entangled feature composed of illumination and reflectance cues ($\mathcal{F}_{img}$), independently. These encodings are reused in the later parts of the architecture, (Fig.~\ref{fig:net_over_view}), allowing independent feature usage for both global and local layers.

\begin{figure}
    \centering
    \includegraphics[width=0.8\linewidth]{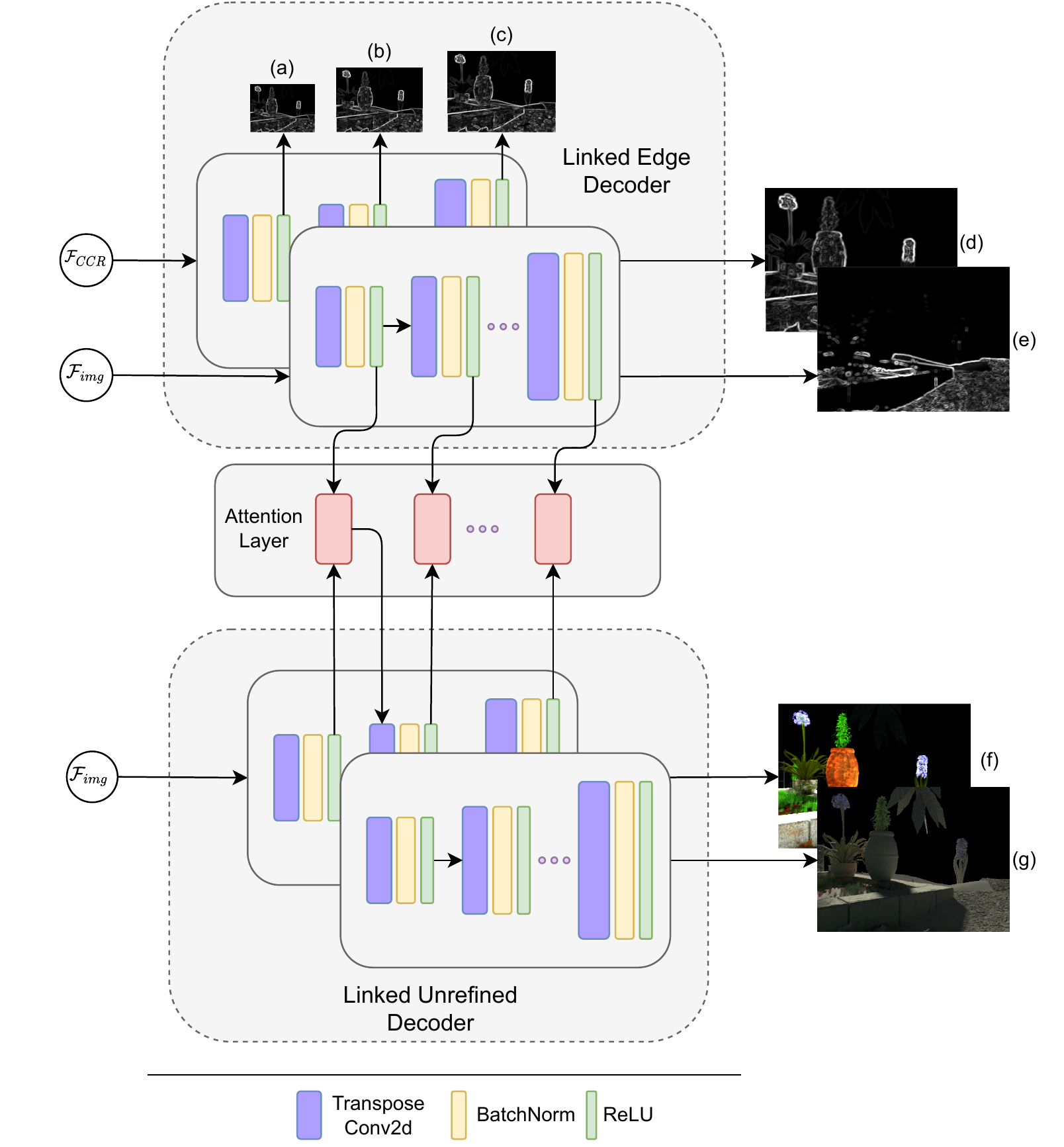}
    \caption{Overview of the linked edge decoder and the linked unrefined decoder. The encoded features from the CCR and the image are used to decode the reflectance (d) and illumination edges (e). Scale space side outputs (a) - (c), used to enforce scale consistency, are also added. The edge features are passed on through the attention layers to the unrefined decoder, which outputs the globally consistent unrefined reflectance (f) and shading (g).}
    \label{fig:linked_module}
\end{figure}

\paragraph{Linked Edge Decoder:}$\mathcal{F}_{CCR}$ \& $\mathcal{F}_{img}$, are passed on to the linked edge decoders (Fig.~\ref{fig:linked_module}). Interconnections within the decoders enable to learn a relational representation of the cues. Thus, the decoder can learn both reflectance edges (d) and illumination edges (e) jointly. Apart from the standard supervision on the output, a feature space scale supervision~\cite{Xie2015} is also added. To facilitate this, two scales ($64\times64$ and $128\times128$) are obtained. These scales are transformed directly from the intermediate CNN features through a common convolution. This allows the convolution to learn a transformation from feature to image space. This supervision ensures that the decoder produces edges that are consistent across scales and feature spaces.

\paragraph{Unrefined Decoder:}The unrefined decoder consists of a similar set of decoders as the edge decoder set as illustrated in Fig.~\ref{fig:linked_module}. For every block in the decoder of the Linked edge decoder, the output is fed through an attention layer before being convolved through the respective block in the unrefined decoder. Skip connections (not shown in the figure, for brevity), are also added to the decoders of $\mathcal{F}_{img}$ to provide additional cues. The attention enhanced edge guidance allows the network to focus on global consistencies for the intrinsic images, like segment wise reflectance consistency and smooth gradient changes for illumination. However, these intrinsics are not necessarily locally consistent, but may still contain local imperfections such as shading-reflectance leakage transitions caused by hard negative illumination transitions. 

\begin{figure}
    \centering
    \includegraphics[width=\linewidth]{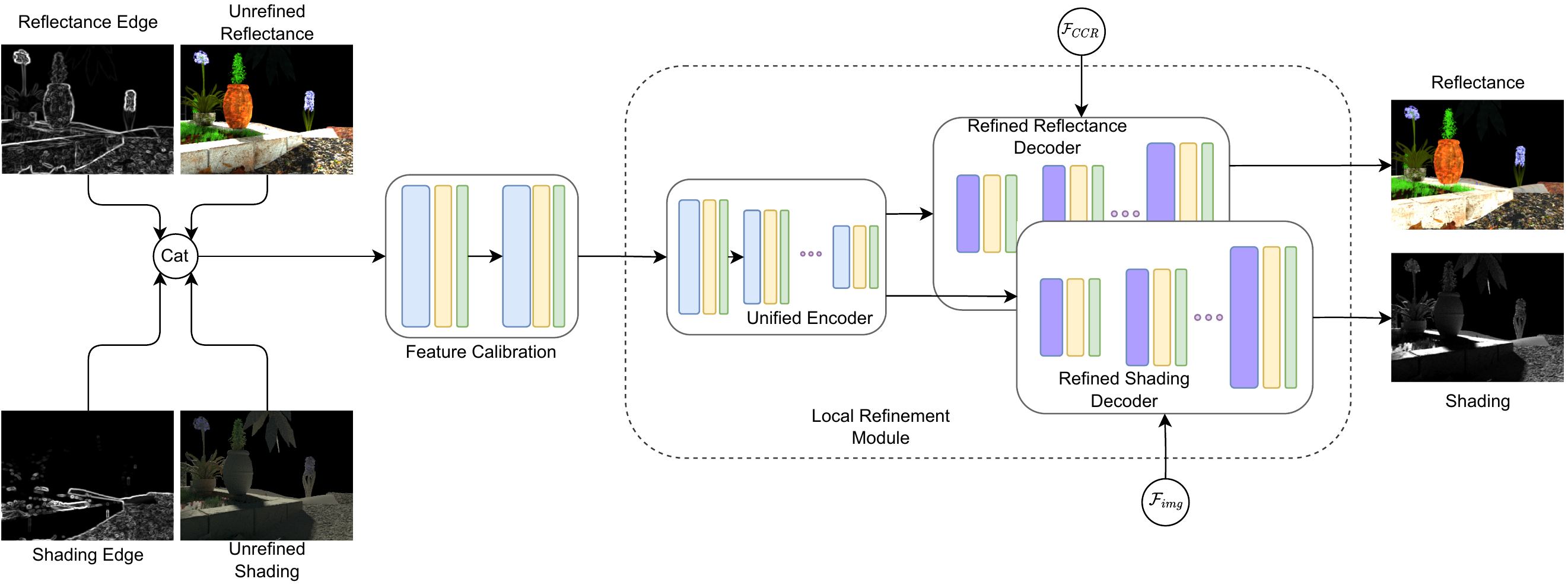}
    \caption{Overview of the local refinement module. The supplementary material provides a higher resolution version.}
    \label{fig:refinement}
\end{figure}

\paragraph{Local Refinement Module:} To cope with hard illumination negatives, the output from the previous decoders are passed on to the refinement module. Fig.~\ref{fig:refinement} illustrates this strategy. The reflectance and shading edges, unrefined reflectance and shading pairs are concatenated and convolved through a feature calibration layer. The calibrated features are then passed through an encoder-decoder to obtain the final output. Additional local patch-wise guidance is provided through skip connections from $\mathcal{F}_{CCR}$ \& $\mathcal{F}_{img}$, which are also passed through attention layers to selectively focus on hard negative areas (not shown in the figure for brevity).

The shading is computed through a separate decoder. The proposed configuration allows the decoder to use the shading cues as an additional source of information to correct the reflectance and vice versa.

\subsection{Loss Functions}

To train the network, supervision is added to each of the output channels of the network. These are: 1) the edge loss ($\mathcal{L}_e$), 2) the unrefined loss ($\mathcal{L}_u$), and 3) the refined loss ($\mathcal{L}_r$). For each of these outputs, a combination of scale invariant MSE~\cite{Narihia2015} and standard MSE loss is used. 

\paragraph{1) Edge Losses:}The edge decoder outputs reflectance and shading edges, together with their scaled versions of $64\times64$ and $128\times128$. In total, there are $3$ outputs for reflectance and shading. The total edge loss is defined by:

\begin{equation}
    \begin{aligned}
        \mathcal{L}_e &=\;\;\;\;\mathcal{L}_{AE}\;+\;\mathcal{L}_{AE64}\;+\;\mathcal{L}_{AE128}\\
                    &\;\;\;+\;\mathcal{L}_{SE}\;+\;\mathcal{L}_{SE64}\;+\;\mathcal{L}_{SE128},
    \end{aligned}
\end{equation}

\noindent where $\mathcal{L}_{AE}$ \& $\mathcal{L}_{SE}$ are the losses on the full scale of reflectance and shading edges; $\mathcal{L}_{AE64}$ \& $\mathcal{L}_{SE64}$ are the losses on reflectance and shading feature outputs at scale $64\times64$; $\mathcal{L}_{AE128}$ \& $\mathcal{L}_{SE128}$ are the losses on the $128\times128$ scale. The ground truth for the edges is calculated using a Canny Edge operator. The reflectance edge is calculated from the reflectance ground truth. The NED~\cite{Baslamisli2018ECCV} dataset (described in more detail in the experimental section) provides fine grained shading decompositions (shadow maps and ambient/inter-reflections). The shadow map is used for the shading edge calculation. For datasets without such ground truth decompositions, the shadow edges are simulated by subtracting the reflectance edges from the shading ground truth edges. Under the Lambertian model, it can be assumed that an image is the multiplication of reflectance and shading. Hence, the subtraction provides an approximation for shading edges.

\paragraph{2) Unrefined Losses:}The unrefined decoder is constrained by the following loss:

\begin{equation}
    \begin{aligned}
        \mathcal{L}_u &=\;\mathcal{L}_{uA}\;+\;\mathcal{L}_{uS},
    \end{aligned}
\end{equation}

\noindent where $\mathcal{L}_{uA}$ is the loss on the unrefined decoder's reflectance output and $\mathcal{L}_{uS}$ on the unrefined decoder's shading output.

\paragraph{3) Refined Loss:}Finally, the following loss is applied on the outputs generated by the refined decoder:

\begin{equation}
    \begin{aligned}
        \mathcal{L}_r &=\;\mathcal{L}_{A}\;+\;\mathcal{L}_{S},
    \end{aligned}
\end{equation}

\noindent where $\mathcal{L}_{A}$ is the loss on the final reflectance output and $\mathcal{L}_{S}$ on the final shading output.

To enforce component dependence as a supervision, the outputs from the network are recombined and compared with the input image for the reconstruction loss $\mathcal{L}_{rec}$. Since this decoder focuses on localised correction of the outputs, a Structural Dissimilarity (DSSIM) loss is added to regularise the network:

\begin{equation}
    \begin{aligned}
        \mathcal{L}_{dssim} &=\;\mathcal{L}_{\delta A}\;+\;\mathcal{L}_{\delta S},
    \end{aligned}
\end{equation}

\noindent where $\mathcal{L}_{\delta A}$ and $\mathcal{L}_{\delta S}$ are the losses on the dissimilarity measure of the reflectance and shading respectively. DSSIM measures the divergence of structural changes. This allows the final reflectance and shading, after both global and local corrections, to be closer to the ground truth.

Finally, to make the network explicitly focus on the perceived quality of the decomposition, a perceptual loss is added. The features of a VGG16~\cite{Simonyan2015} network trained on ImageNet are used. This is defined as follows:

\begin{equation}
    \mathcal{L}_P (A, \hat{A}) = \sum_i || \mathcal{F}_{VGG_i}(A) - \mathcal{F}_{VGG_i}(\hat{A}) ||_1.
\end{equation}

\noindent where $\mathcal{L}_P$ is the perceptual loss; $\mathcal{F}_{VGG}$ is the feature space transform function; $A$ is the predicted reflectance; $\hat{A}$ is the corresponding ground truth; $i$ is the layer index of the VGG16 network (set to the last $4$ for all the experiments).

\noindent Combining all the losses, the total training objective for the network is defined by:

\begin{equation}
\begin{aligned}
    \mathcal{L} = \mathcal{L}_{r} + \lambda_u \; \mathcal{L}_{u}
    + \lambda_e \; \mathcal{L}_{e} + \lambda_p \; \mathcal{L}_p \\
    + \lambda_d \; \mathcal{L}_{dssim} + \mathcal{L}_{rec}.
\end{aligned}
\end{equation}

\noindent where, the component specific hyper-parameters $\lambda_u$, $\lambda_d$, $\lambda_e$ and $\lambda_p$ are empirically set to be $0.5$, $0.4$, $0.4$ and $0.05$ respectively. For more details please see the supplementary material.

\section{Experiments and Results}

\paragraph{Datasets} Quantitative experiments are conducted on four datasets:  NED~\cite{Baslamisli2018ECCV}, MIT~\cite{Grosse2009}, Sintel~\cite{Butler2012} and IIW~\cite{Bell2014}. The train/test splits, as provided by the original papers, are used. The proposed network is trained on the NED dataset and finetuned on the other datasets. In addition, qualitative results are provided on the Trimbot dataset~\cite{Sattler2017}, which consists of real-world garden dataset. This dataset does not come with ground truth annotations.

\paragraph{Evaluation Metric.} Following the literature, the standard MSE error metric, the LMSE metric~\cite{Grosse2009}, (with a windows size of 20) and the structural dissimilarity metric (DSSIM) are used. Finally, for IIW, the WHDR metric~\cite{Bell2014} is used. For all the datasets, the same train and test splits are used for all methods.

\subsection{Ablation Study}

\paragraph{Influence of Illumination Invariants:} In this experiment, the influence of the CCR to steer the global-local process is studied by removing the CCR encoder from the network. In this way, the edge decoder is completely dependent on the $RGB$ image cues given by the input. ~\cref{tab:ablation_architecture} shows the results of this experiment.

\begin{table}[ht]
\centering

\resizebox{0.48\textwidth}{!}{%
\begin{tabular}{c|c|c|c|c|c|c|}
\cline{2-7}
                                         & \multicolumn{3}{c|}{Reflectance} & \multicolumn{3}{c|}{Shading} \\ \cline{2-7} 
                                         & MSE    & LMSE   & DSSIM  & MSE    & LMSE   & DSSIM  \\ \hline
\multicolumn{1}{|c|}{w/o Physics Priors} & 0.0039 & 0.0449 & 0.2590 & 0.0032 & 0.0812 & 0.2356 \\ \hline
\multicolumn{1}{|c|}{w/o Edge Guidance} & 0.0033 & 0.0415 & 0.2411 & 0.0029 & 0.0782 & 0.2543 \\ \hline
\multicolumn{1}{|c|}{w Canny Edge Guidance} & 0.0061 & 0.0566 & 0.2721 & 0.0034 & 0.0879 & 0.2538 \\ \hline
\multicolumn{1}{|c|}{w/o Local Refinement} & 0.0020 & 0.0361 & 0.1192 & 0.0031 & 0.0768 & 0.2651 \\ \hline
\multicolumn{1}{|c|}{w/o Attention Layers} & 0.0019    & 0.0330    & 0.0776   & 0.0026   & 0.0704  & 0.1301  \\ \hline
\multicolumn{1}{|c|}{Proposed}           & \textbf{0.0015} & \textbf{0.0289} & \textbf{0.0688} & \textbf{0.0018} & \textbf{0.0489} & \textbf{0.1005} \\ \hline
\end{tabular}%
}
\caption{An ablation study on the various parts of our network. From the results, the proposed component does indeed have a positive influence on the performance of the network. w - with, w/o - without}
\label{tab:ablation_architecture}
\end{table}

From~\cref{tab:ablation_architecture}, it is shown that the removal of the CCR (\textit{w/o Physics Priors}) degrades the performance of the proposed network. This is because the modified network now only relies on the $RGB$ edges of the input image. These edges include strong illumination transitions and therefore the network is sensitive to hard negatives. As a result, all the metrics show a decrease in performance. In conclusion, it is beneficial for edge-driven hybrid IID networks to make use of illumination invariant descriptors, rather than learning a data-distribution. 

\paragraph{Influence of Reflectance and Shading Edges:} (1) The influence of edges as a source of guidance is analysed. The edge decoder part is removed from the unrefined decoder. The CCR features are maintained in the local refinement module. (2) It is tested whether learning the reflectance and shading edges computed directly from the image edges can replace the CCR to the edge translation subnet. Therefore, the input to the CCR encoder in Fig.~\ref{fig:net_over_view} is replaced by Canny edges calculated directly from the images. This setup corresponds to~\cite{Fan2018}.

For the setup described in $(1)$, results for removing the edge guidance, (\textit{w/o Edge Guidance}) in~\cref{tab:ablation_architecture} shows that the performance decreases. However, it is still an improvement over the previous experiment (\textit{w/o Physics Priors}) in the table. This shows that even if the global edge guidance is removed, the local physics prior is still able to help. 

In setup $(2)$, using only Canny edges is shown to degrade the performance. This is because the edge-to-edge transition is lacking any guidance. This subsequently makes the local correction fail. Thus, this experiment shows the importance of consistent global guidance for intrinsic image decomposition.

\paragraph{Influence of Local Refinement:} The last refinement module is removed and only the edge guided module is kept. This makes the decoder module to handle both global and local consistencies in the same parameter space. The result for this experiment is shown in~\cref{tab:ablation_architecture}. The results show that there is an improvement for the explicit global \& local parameter disentanglement. This shows that the global and local context separation is an integral part of the proposed network architecture.

\paragraph{Influence of Attention Layers:} In this experiment, all the attention layers are removed from the network and replaced by direct connections. The result of this experiment is also shown in~\cref{tab:ablation_architecture}.

The results (\textit{w/o Attention Layers} compared to \textit{Proposed}) show that the inclusion of the attention mechanism improves the performance. The metrics (LMSE \& DSSIM) demonstrate improvements indicating that the attention layers are beneficial for local corrections. This means that there is no single uniform transformation to be applied to all pixels in an image to recover the intrinsic components. A detailed study supporting this hypothesis can be found in the supplementary material.

\subsection{Evaluations \& Results}

\begin{figure}
    \centering
    \includegraphics[width=\linewidth,height=0.35\linewidth]{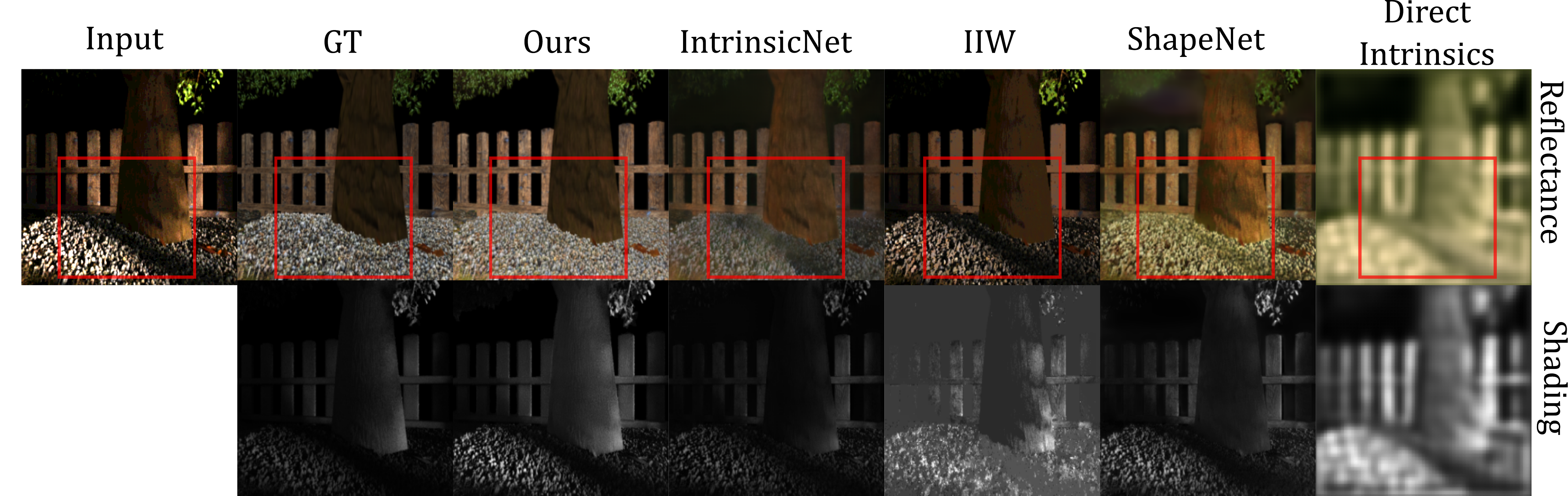}
    \caption{Comparison of the proposed network with sota methods. In the image, the tree trunk has different textures and has a hard negative illumination transition on the base. It is shown that the proposed method can both recover from hard negatives and also prevent shadow-reflectance misclassifications in the shading component.}
    \label{fig:sota_figure}
\end{figure}

\paragraph{Comparison on NED Dataset:} In this experiment, the proposed network is compared to state-of-the-art (sota) methods. All presented methods are re-trained using the NED dataset. For a fair comparison, the same train and test split are used as well as the optimum parameters as mentioned in the respective papers. Recent unsupervised methods of~\cite{Liu2020},~\cite{Li2018CVPR} and~\cite{Yu2019} are also included for comparison. The numerical results are shown in~\cref{tab:sota_compare} and the visual results in Fig.~\ref{fig:sota_figure}. 

\begin{table}[ht]
\centering
\resizebox{0.48\textwidth}{!}{%
\begin{tabular}{c|cccccc|}
\cline{2-7}
 & \multicolumn{3}{c|}{Reflectance} & \multicolumn{3}{c|}{Shading} \\ \cline{2-7} 
 & \multicolumn{1}{c|}{MSE} & \multicolumn{1}{c|}{LMSE} & \multicolumn{1}{c|}{DSSIM} & \multicolumn{1}{c|}{MSE} & \multicolumn{1}{c|}{LMSE} & DSSIM \\ \cline{2-7} \cline{2-7}
\multicolumn{1}{l|}{} & \multicolumn{6}{c|}{Supervised methods} \\ \hline
\multicolumn{1}{|c|}{Color Retinex~\cite{Grosse2009}} & \multicolumn{1}{c|}{0.0114} & \multicolumn{1}{c|}{0.1204} & \multicolumn{1}{c|}{0.3280} & \multicolumn{1}{c|}{0.0193} & \multicolumn{1}{c|}{0.2334} & 0.3515 \\ \hline
\multicolumn{1}{|c|}{IIW~\cite{Bell2014}} & \multicolumn{1}{c|}{0.0095} & \multicolumn{1}{c|}{0.1343} & \multicolumn{1}{c|}{0.2098} & \multicolumn{1}{c|}{0.0111} & \multicolumn{1}{c|}{0.1861} & 0.3511 \\ \hline
\multicolumn{1}{|c|}{Direct Intrinsics~\cite{Narihia2015}} & \multicolumn{1}{c|}{0.0073} & \multicolumn{1}{c|}{0.1205} & \multicolumn{1}{c|}{0.3756} & \multicolumn{1}{c|}{0.0065} & \multicolumn{1}{c|}{0.1798} & 0.3843 \\ \hline
\multicolumn{1}{|c|}{IntrinsicNet~\cite{Baslamisli2018CVPR}} & \multicolumn{1}{c|}{0.0035} & \multicolumn{1}{c|}{0.0449} & \multicolumn{1}{c|}{0.2367} & \multicolumn{1}{c|}{0.0037} & \multicolumn{1}{c|}{0.0791} & 0.2110 \\ \hline
\multicolumn{1}{|c|}{ShapeNet~\cite{Shi2017}} & \multicolumn{1}{c|}{0.0053} & \multicolumn{1}{c|}{0.0597} & \multicolumn{1}{c|}{0.2516} & \multicolumn{1}{c|}{0.0050} & \multicolumn{1}{c|}{0.0910} & 0.2186 \\ \hline \cline{2-7}
\multicolumn{1}{l|}{} & \multicolumn{6}{c|}{Unsupervised methods} \\ \hline
\multicolumn{1}{|c|}{USI3D~~\cite{Liu2020}} & \multicolumn{1}{c|}{0.0081} & \multicolumn{1}{c|}{0.0360} & \multicolumn{1}{c|}{0.1886} & \multicolumn{1}{c|}{0.0143} & \multicolumn{1}{c|}{0.0608} & 0.2140 \\ \hline
\multicolumn{1}{|c|}{IIDWW~~\cite{Li2018CVPR}} & \multicolumn{1}{c|}{0.0149} & \multicolumn{1}{c|}{0.0447} & \multicolumn{1}{c|}{0.2229} & \multicolumn{1}{c|}{0.0175} & \multicolumn{1}{c|}{0.0698} & 0.2346 \\ \hline
\multicolumn{1}{|c|}{InverseRenderNet~~\cite{Yu2019}} & \multicolumn{1}{c|}{0.0478} & \multicolumn{1}{c|}{0.0642} & \multicolumn{1}{c|}{0.2751} & \multicolumn{1}{c|}{0.0505} & \multicolumn{1}{c|}{0.2597} & 0.3382 \\ \hline \hline
\multicolumn{1}{|c|}{Ours} & \multicolumn{1}{c|}{\textbf{0.0015}} & \multicolumn{1}{c|}{\textbf{0.0289}} & \multicolumn{1}{c|}{\textbf{0.0688}} & \multicolumn{1}{c|}{\textbf{0.0018}} & \multicolumn{1}{c|}{\textbf{0.0489}} & \textbf{0.1005} \\ \hline
\end{tabular}%
}
\caption{Numerical evaluation comparison between the proposed architecture and sota baselines on the NED dataset. The first group are supervised methods, second group are unsupervised methods and the final group is the our proposed methods.}
\label{tab:sota_compare}
\end{table}

The results show that the proposed method outperforms the baselines for all metrics. As illustrated in Fig.~\ref{fig:sota_figure}, the proposed method recovers the intrinsics more robustly. For example, the proposed method is able to disentangle the shadows at the base of the tree, while other methods suffer from hard negatives resulting in discoloured reflectances. 

\begin{figure}[ht]
    \centering
    \includegraphics[width=\linewidth,height=0.7\linewidth]{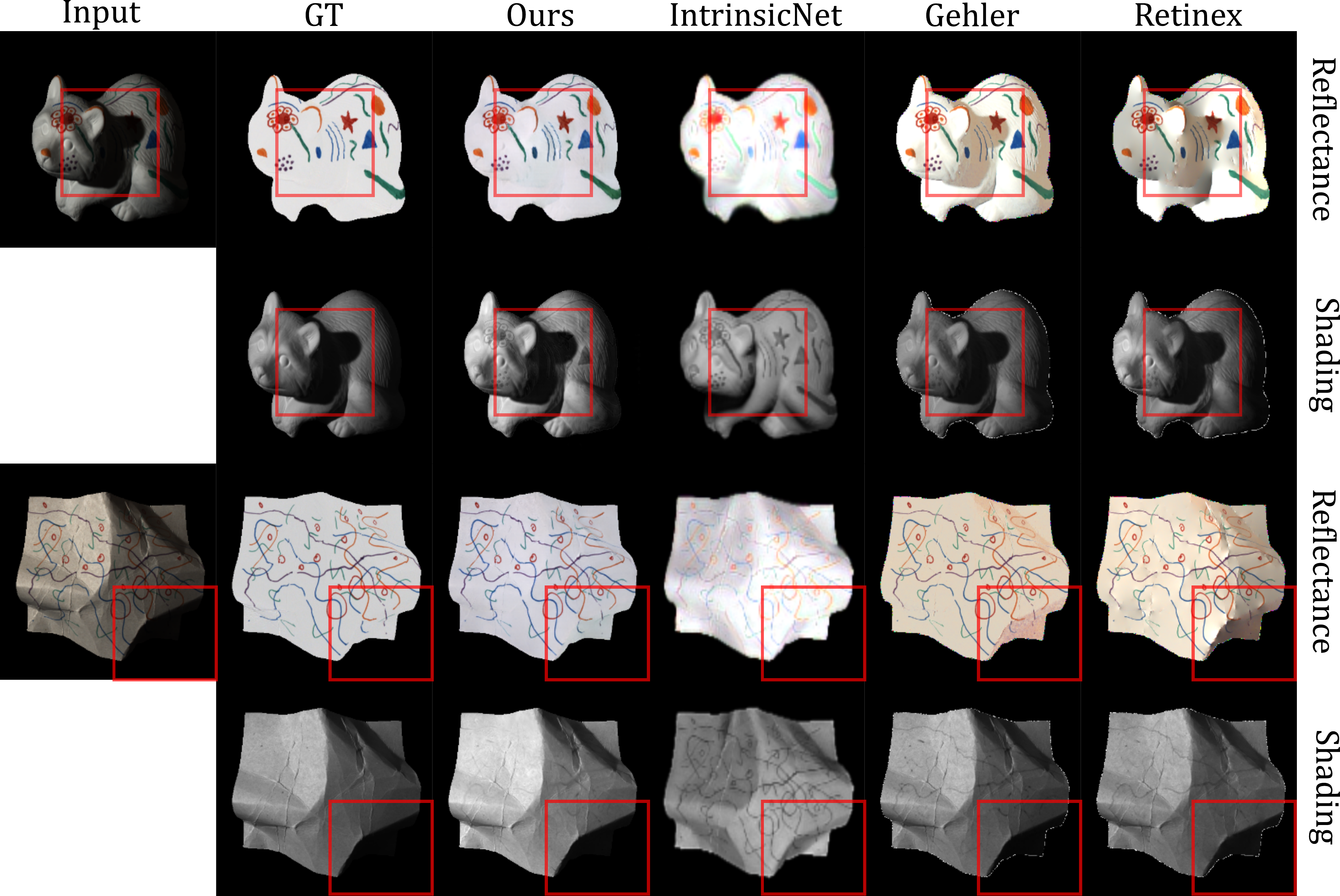}
    \caption{Qualitative evaluation of the proposed network on the MIT dataset. The proposed method is the only method able to disentangle shadows from reflectance cues.}
    \label{fig:mit_figure}
\end{figure}

\paragraph{MIT Intrinsics Dataset:} The proposed network is finetuned on the MIT Intrinsics dataset. The quantitative numbers are shown in~\cref{tab:mit_compare}, while visuals are shown in Fig.~\ref{fig:mit_figure}

\begin{table}[ht]
\centering
\resizebox{0.48\textwidth}{!}{%
\begin{tabular}{c|c|c|c|c|c|c|}
\cline{2-7}
                                         & \multicolumn{3}{c|}{Reflectance} & \multicolumn{3}{c|}{Shading} \\ \cline{2-7} 
                                         & MSE       & LMSE      & DSSIM    & MSE      & LMSE    & DSSIM   \\ \cline{2-7} \cline{2-7}
\multicolumn{1}{l|}{} & \multicolumn{6}{c|}{Supervised methods} \\ \hline
\multicolumn{1}{|c|}{SIRFS~\cite{Barron2015}} & 0.0129    & 0.0572    & -        & 0.0066   & 0.0309  & -       \\ \hline
\multicolumn{1}{|c|}{Gehler~\etal~\cite{Gehler2011}} & 0.0065    & 0.0393    & -        & 0.0051   & 0.0282  & -       \\ \hline
\multicolumn{1}{|c|}{Zhou~\etal~\cite{Zhou2015}}   & 0.0252    & -         & -        & 0.0229   & -       & -       \\ \hline
\multicolumn{1}{|c|}{Color Retinex~\cite{Grosse2009}} &
  0.0084 &
  0.0447 &
  - &
  0.076 &
  0.0343 &
  - \\ \hline
\multicolumn{1}{|c|}{Direct Intrinsics~\cite{Narihia2015}} &
  0.0277 &
  0.0585 &
  0.1526 &
  0.0154 &
  0.0295 &
  0.1328 \\ \hline
\multicolumn{1}{|c|}{ShapeNet~\cite{Shi2017}}           & 0.0278    & 0.0503    & 0.1465   & 0.0126   & 0.0240  & 0.1200  \\ \hline
\multicolumn{1}{|c|}{CGIntrinsics~\cite{Li2018ECCV}} & 0.0221    & 0.0349    & 0.1739   & 0.0186   & 0.0259  & 0.1652  \\ \hline
\multicolumn{1}{|c|}{\begin{tabular}[c]{@{}c@{}}CGIntrinsics~\cite{Li2018ECCV} \\ (MIT Finetuned)\end{tabular}}
& 0.0167    & 0.0319    & 0.1287   & 0.0127   & 0.0211  & 0.1376  \\ \hline
\multicolumn{1}{|c|}{ParCNN~\cite{Yuan2019}}           & 0.0109 & 0.0462 & 0.0929 & 0.0086 & 0.0537 & 0.0999 \\ \hline
\multicolumn{1}{|c|}{CasQNet~\cite{Ma2020}}       & 0.0091    & 0.0212    & 0.0730   & 0.0081   & 0.0192  & 0.0659  \\ \hline
\multicolumn{1}{|c|}{IntrinsicNet~\cite{Baslamisli2018CVPR}} & 0.0104 & 0.0854 & - &   0.0304 & 0.2038 & - \\ \hline
\multicolumn{1}{|c|}{Baslamisli~\etal~\cite{Baslamisli2020}} & 0.0060 & 0.0438 & - & 0.0069 & 0.0418 & - \\ \hline \cline{2-7}
\multicolumn{1}{l|}{} & \multicolumn{6}{c|}{Unsupervised methods} \\ \hline
\multicolumn{1}{|c|}{STAR~\cite{Xu2020}}             & 0.0137 & 0.0614 & 0.1196 & 0.0114 & 0.0672 & 0.0825 \\ \hline 
\multicolumn{1}{|c|}{USI3D~\cite{Liu2020}}            & 0.0156 & 0.0640 & 0.1158 & 0.0102 & 0.0474 & 0.1310 \\ \hline 
\multicolumn{1}{|c|}{IIDWW~\cite{Li2018ECCV}}            & 0.0126 & 0.0591 & 0.1049 & 0.0105 & 0.0457 & 0.1159 \\ \hline 
\multicolumn{1}{|c|}{InverseRenderNet~\cite{Yu2019}} & 0.0234 & 0.0573 & 0.1148 & 0.0186 & 0.0765 & 0.1276 \\\cline{2-7} \hline \hline

\multicolumn{1}{|c|}{Ours} &
  \textbf{0.0028} &
  \textbf{0.0136} &
  \textbf{0.0340} &
  \textbf{0.0035} &
  \textbf{0.0183} &
  \textbf{0.0493} \\ \hline
\end{tabular}%
}
\caption{Quantitative evaluation comparison of the proposed architecture on the MIT Intrinsic Dataset~\cite{Grosse2009}.}
\label{tab:mit_compare}
\end{table}

The results show that the proposed method outperforms other baseline methods for all the metrics. The method is able to recover the intrinsic components robustly. For example, the shading map obtained by IntrinsicNet on the raccoon completely misses the shadow. For our method, the reflectance of the paper is much smoother.

\paragraph{MPI Sintel Dataset:} The results on the MPI Sintel Dataset are given in~\cref{tab:sintel_compare}.

\begin{table}[ht]
\centering
\resizebox{0.48\textwidth}{!}{%
\begin{tabular}{c|c|c|c|c|c|c|}
\cline{2-7}
                                        & \multicolumn{3}{c|}{Reflectance} & \multicolumn{3}{c|}{Shading} \\ \cline{2-7} 
                                        & MSE       & LMSE      & DSSIM    & MSE      & LMSE    & DSSIM   \\ \hline
\multicolumn{1}{|c|}{Color Retinex~\cite{Grosse2009}}     & 0.0606    & 0.0366    & 0.2270   & 0.0727   & 0.0419  & 0.2400  \\ \hline
\multicolumn{1}{|c|}{Lee~\etal~\cite{Lee2012}}   & 0.0463    & 0.0224    & 0.1990   & 0.0507   & 0.0192  & 0.1770  \\ \hline
\multicolumn{1}{|c|}{SIRFS~\cite{Barron2013}}             & 0.0420    & 0.0298    & 0.2100   & 0.0436   & 0.0264  & 0.2060  \\ \hline
\multicolumn{1}{|c|}{Chen~\etal~\cite{Chen2013}}  & 0.0307    & 0.0185    & 0.1960   & 0.0277   & 0.0190  & 0.1650  \\ \hline
\multicolumn{1}{|c|}{Direct Intrinsics~\cite{Narihia2015}} & 0.0100    & 0.0083    & 0.2014   & 0.0092   & 0.0085  & 0.1505  \\ \hline
\multicolumn{1}{|c|}{Fan~\etal~\cite{Fan2018}}   & 0.0069    & \textbf{0.0044}    & 0.1194   & \textbf{0.0059}   & \textbf{0.0042}  & 0.0822  \\ \hline \hline
\multicolumn{1}{|c|}{Ours}          & \textbf{0.0015}    & 0.0080    & \textbf{0.0399}   & 0.0105   & 0.0507  & \textbf{0.0508}  \\ \hline
\end{tabular}%
}
\caption{Standard numerical evaluation comparison of the proposed method on the MPI Sintel Dataset~\cite{Butler2012} (image split).}
\label{tab:sintel_compare}
\end{table}

The proposed method generally outperforms, on average, all other methods except for the LMSE metric. Particularly, the proposed method outperforms other methods for the DSSIM metric for both components and hence robustly preserves global and local structural components. From the outputs (included in the supplementary material), it is shown that the shading computed by the proposed method has a lower pixel scale but is structurally consistent. This explains the lower performance on other metrics compared to the DSSIM metric (for shading). The MSE and LMSE metrics are more sensitive to outliers and optimise to a smaller Euclidean distance. Additional details can be found in the supplementary material. 

\begin{figure}[ht]
    \centering
    \includegraphics[width=0.9\linewidth]{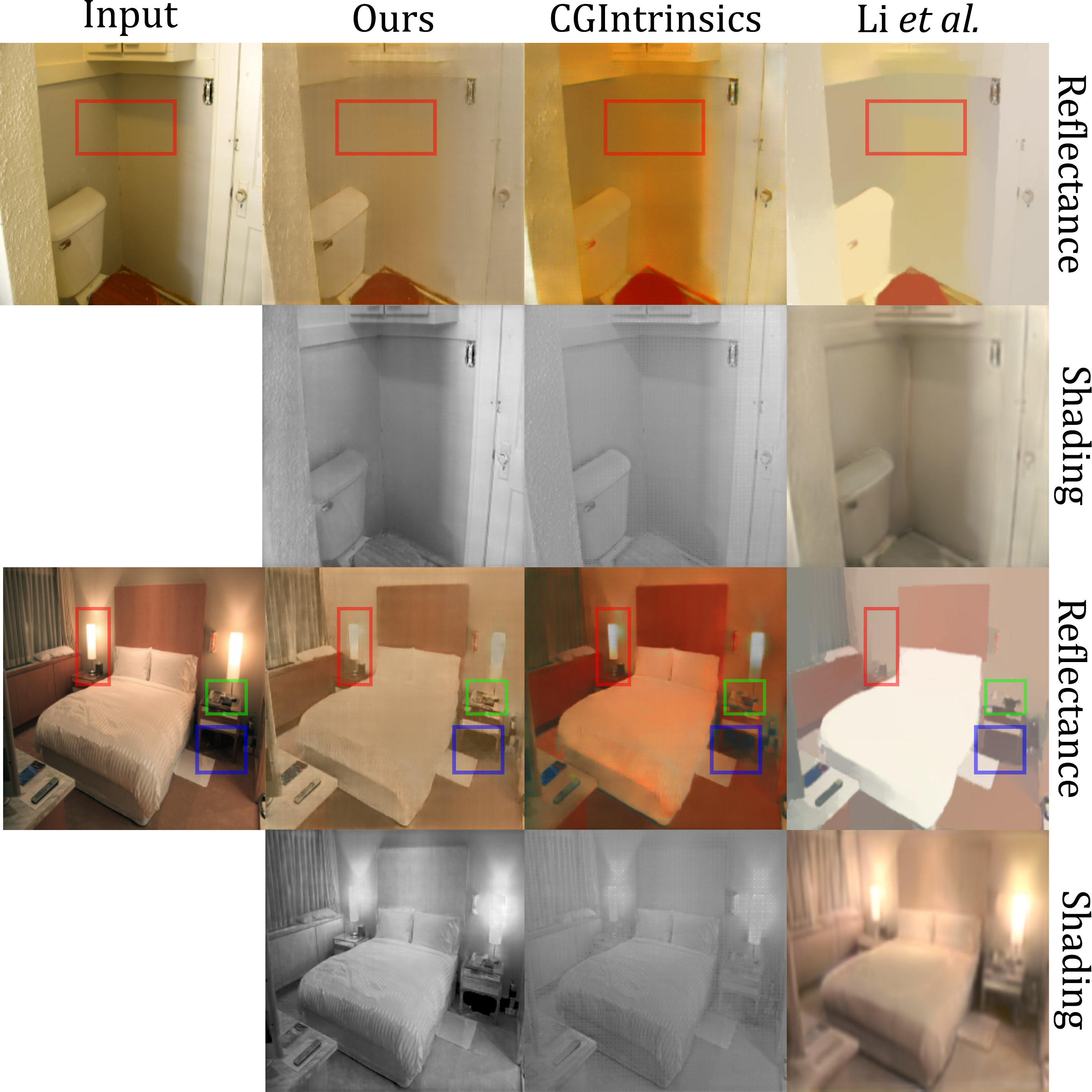}
    \caption{Results of the proposed method compared with CGIntrinsics~\cite{Li2018ECCV} and~\cite{Li2020}. The proposed method shows comparable performance with other methods, despite being trained primarily on outdoor garden images.}
    \label{fig:iiw_figure}
\end{figure}

\paragraph{IIW Dataset:} For this experiment, the proposed network is finetuned on the IIW dataset. The results for the network are given in~\cref{tab:whdr_comp}. Visuals are shown in Fig.~\ref{fig:iiw_figure}.

\begin{table}[ht]
\centering
\resizebox{0.4\textwidth}{!}{%
\begin{tabular}{|c|c|c|}
\hline
Methods           & WHDR (mean)    & WHDR (Outdoors only) \\ \hline
Direct Intrinsics~\cite{Narihia2015} & 37.3          & -                    \\ \hline
Color Retinex~\cite{Grosse2009}    & 26.9          & -                    \\ \hline
Garces~\etal~\cite{Garces2012}            & 25.5          & -                    \\ \hline
Zhao~\etal~\cite{Zhao2012}              & 23.2          & -                    \\ \hline
IIW~\cite{Bell2014}               & 20.6          & 21.7                \\ \hline
Nestmeyer~\etal~\cite{Nestmeyer2016}     & 19.5          & -                    \\ \hline
Bi~\etal~\cite{Bi2015}                & 17.7          & -                    \\ \hline
Sengupta~\etal~\cite{Sengupta2019} & 16.7 & - \\ \hline
Li~\etal~\cite{Li2020} & 15.9 & 21.3 \\ \hline
CGIntrinsics~\cite{Li2018ECCV}      & 15.5 & 23.1                \\ \hline
GLoSH~\cite{Zhou2019} & 15.2 & - \\ \hline 
Fan~\etal~\cite{Fan2018} & 15.4 & 21.6 \\ \hline
Ours          & 21.3          & 20.8       \\ \hline \hline
Fan~\etal*~\cite{Fan2018} & \textbf{14.45} & 20.2 \\ \hline
Our* & 18.5 & \textbf{18.4} \\ \hline
\end{tabular}%
}
\caption{Performance in terms of the WHDR metric. The proposed method is trained with general image learning losses. When testing it only on outdoor images, the proposed method shows competitive performance. * denotes outputs post-processed with a guided filter.}
\label{tab:whdr_comp}
\end{table}

Due to the nature of the ground truth provided by this dataset, the proposed method can only use the ordinal loss~\cite{Bell2014} for finetuning on the train set. The original proposed network is trained on the MSE and perceptual losses only. However, it is still able to perform comparatively. GLoSH~\cite{Zhou2019} is the best performing method. However, it needs both the normal and lighting ground truth for supervision. Fig.~\ref{fig:iiw_figure} shows flatter reflectance cues (on the wall and bed), insensitive to illumination patterns (i.e. shadows and shading). ~\cite{Li2020} needs supervision on normals, depth, roughness and lighting, in addition to the reflectance and shading and has $7$ separate training stages. The reflectance cues are missing details, while the shading cues are quite blurry. The proposed network is trained only on outdoor images (domain difference). To test the domain related performance, the WHDR metric is applied to only the outdoor images in the test set. It is  shown that the proposed method performs best.

\begin{figure}[ht]
    \centering
    \includegraphics[width=0.785\linewidth]{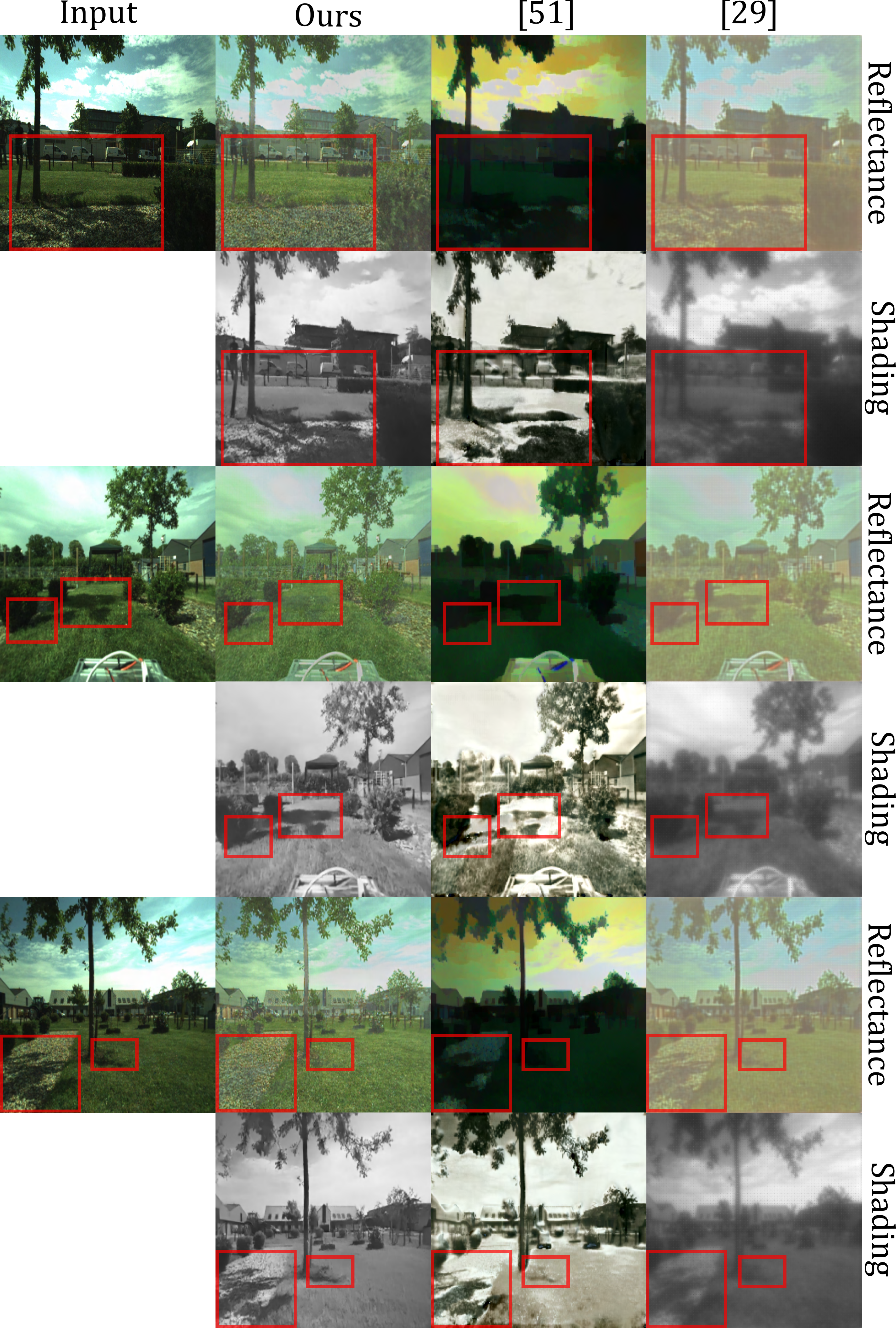}
    \caption{Results of the proposed method on the Trimbot dataset. The proposed method is trained and finetuned on a fully synthetic dataset, yet it can recover proper reflectances by removing both soft and hard illumination patterns. The supplementary materials contain additional results on in-the-wild images.}
    \label{fig:trimbot_figure}
\end{figure}

\paragraph{Trimbot Dataset:} To test the generalisation of the proposed method to real world scenarios, results on the Trimbot Dataset~\cite{Sattler2017} are shown in Fig.~\ref{fig:trimbot_figure}. 

\section{Conclusion}
In this paper, an end-to-end edge-driven hybrid approach has been proposed for intrinsic image decomposition. Edges are based on illumination invariant descriptors. To handle hard negative illumination transitions, a hierarchical approach has been taken including global and local refinement layers. The global guidance was integrated through a reflectance and shading edge formulation. For the local guidance, the encoded CCR features were used as a prior to the local refinement module. 

Based on extensive ablation study and large scale experiments, is has been shown that (1) it is beneficial for edge-driven hybrid IID networks to make use of illumination invariant descriptors, (2) separating global and local cues into different modules indeed helps in improving the performance of the network both qualitatively and quantitatively, (3) the proposed method obtains sota performance in recovering the intrinsics, and (4) it is able to generalise well to real world images.

\section{Appendix A: Image Formation}

In this section we briefly introduce the image formation model. We follow this with the physics-based descriptors.

\subsection{Image Formation}
\label{sec:imf}
The process of image formation can be modelled as the combination of an object's material and geometric property under a light source. Under the Lambertian assumption~\cite{Shafer1985}, it is:
\begin{equation} \label{eq:imf_suppl}
\begin{split}
I = m(\vec{n}, \vec{l}) \int_{\omega}^{} \rho_{b}(\lambda)\; e(\lambda)\;f(\lambda)\; \mathrm{d}\lambda \;,
\end{split}
\end{equation}

\noindent where, $I$ is the final observed image. $\lambda$ is the incoming wavelength of the light integrated over the entire visible spectrum $\omega$. $\vec{n}$ denotes the surface normal while $\vec{l}$ denotes the incident light direction. $m$ is a function denoting the interaction of $\vec{n}$ \& $\vec{l}$. $f$ indicates the spectral camera sensitivity. $e$ describes the spectral power distribution of the illuminant. While $\rho$ denotes the reflectance of the object.

Discretising the model, we obtain:

\begin{equation}
    \begin{aligned}
        C_{p_1} = m(\vec{n},\;\vec{l})\;e^{C_{p_1}}(\lambda)\;
        \rho^{C_{p_1}}(\lambda)\;,
    \end{aligned}{}\label{eq:chan_eq_suppl}
\end{equation}{}

\noindent where $C_{p_1}$ is colour channel $C$ for pixel $p_1$ for a $RGB$ image.

Assuming a white light source, a narrow band filter and linear sensor response of the device, Eq~\eqref{eq:chan_eq_suppl} can be further simplified as:

\begin{equation} \label{eq:iid_suppl}
I = S \times R \;,
\end{equation}

\noindent where the $m(\vec{n},\;\vec{l})$ component is denoted as $S$ and $e^{C_{p_1}}(\lambda)$ \& $\rho^{C_{p_1}}(\lambda)$ components as $R$, respectively. Consecutively, $S$ is the shading component associated with only the geometry and illuminant of the scene. While $R$ is the reflectance (albedo) image, corresponding to the (true) colour of the object, independent of any geometric or lighting information. Thus, the simplified image formation becomes the pixel-wise multiplication of the reflectance and the shading images.

\subsection{Cross Color Ratios: Proof of Invariance}

The detailed steps of deriving the CCR is shown. Given the CCR ($M_{RG}$ for the channel $RG$ pair for pixels $p_1$ and $p_2$:

\begin{equation} \label{eq:cross_colour_ratios_suppl}
\begin{aligned}
        M_{RG} = \;\frac{R_{p_1}\;G_{p_2}}{R_{p_2}\;G_{p_1}}\;,
\end{aligned}
\end{equation}

\noindent Taking logarithm on both sides of the equation, we get:

\begin{equation} \label{eq:cross_colour_ratios_2_suppl}
\begin{aligned}
        log(M_{RG}) &= \;log(R_{p_1}\;G_{p_2})\; - \;log(R_{p_2}\;G_{p_1})\;,\\
\end{aligned}
\end{equation}

Combining Eq~\eqref{eq:chan_eq_suppl} and Eq~\eqref{eq:cross_colour_ratios_2_suppl}:

\begin{equation} \label{eq:M_sanity_suppl}
\begin{aligned}
        log(M_{RG}) &=\;\;\;\;log(R_{p_1}\;G_{p_2})\;-\;log(R_{p_2}\;G_{p_1})\;,\\
        log(M_{RG}) &=\;\;\;\;log(R_{p_1}) \;+\; log(G_{p_2})\;\\
          &\;\;\;-\;log(R_{p_2})\;-\; log(G_{p_1})\;,\\
        log(M_{RG}) &=\;\;\;\;log(m(\vec{n_{p_1}}\;\vec{l_{p_1}}))\;+\; log(e^{R_{p_1}}(\lambda))\;\\
        &\;\;\;+\;log(\rho^{R_{p_1}}(\lambda))\;+\;log(m(\vec{n_{p_2}}\;\vec{l_{p_2}}))\;\\&\;\;\;+\; log(e^{G_{p_2}}(\lambda))\; +\;log(\rho^{G_{p_2}}(\lambda))\;\\
          &\;\;\;-\;log(m(\vec{n_{p_2}}\;\vec{l_{p_2}}))\;-\; log(e^{R_{p_2}}(\lambda))\;\\                  &\;\;\;-\;log(\rho^{R_{p_2}}(\lambda))\;-\;log(m(\vec{n_{p_1}}\;\vec{l_{p_1}}))\;\\
          &\;\;\;-\; log(e^{G_{p_1}}(\lambda))\; -\;log(\rho^{G_{p_1}}(\lambda))\;,\\
\end{aligned}
\end{equation}

Recall from the main paper:

\begin{equation} \label{eq:illum_eq_suppl}
    e^{C_{p_1}} = e^{C_{p_2}}\;,
\end{equation}

Even for a curved surfaces, this holds true, since they are very close to each other. However, for curved surfaces, the geometry might not be the same, i.e.:

\begin{equation}
    \begin{aligned}
        \vec{n_{p_1}}\;\neq\;\vec{n_{p_2}}\;,
    \end{aligned}
\end{equation}

Incorporating Eq~\eqref{eq:illum_eq_suppl} in Eq~\eqref{eq:M_sanity_suppl}, we have:

\begin{equation} \label{eq:CCR_step1_suppl}
\begin{aligned}
        log(M_{RG}) &=\;\;\;\;log(m(\vec{n_{p_1}}\;\vec{l_{p_1}}))\;+\; log(e^{R_{p_1}}(\lambda))\;\\
        &\;\;\;+\;log(\rho^{R_{p_1}}(\lambda))\;+\;log(m(\vec{n_{p_2}}\;\vec{l_{p_2}}))\;\\&\;\;\;+\; log(e^{G_{p_2}}(\lambda))\; +\;log(\rho^{G_{p_2}}(\lambda))\;\\
          &\;\;\;-\;log(m(\vec{n_{p_2}}\;\vec{l_{p_2}}))\;-\; log(e^{R_{p_2}}(\lambda))\;\\                  &\;\;\;-\;log(\rho^{R_{p_2}}(\lambda))\;-\;log(m(\vec{n_{p_1}}\;\vec{l_{p_1}}))\;\\
          &\;\;\;-\; log(e^{G_{p_1}}(\lambda))\; -\;log(\rho^{G_{p_1}}(\lambda))\;,\\
        log(M_{RG}) &=\;\;\;\;log(m(\vec{n_{p_1}}\;\vec{l_{p_1}}))\;+\; log(e^{R_{p_2}}(\lambda))\;\\
        &\;\;\;+\;log(\rho^{R_{p_1}}(\lambda))\;+\;log(m(\vec{n_{p_2}}\;\vec{l_{p_2}}))\;\\&\;\;\;+\; log(e^{G_{p_2}}(\lambda))\; +\;log(\rho^{G_{p_2}}(\lambda))\;\\
          &\;\;\;-\;log(m(\vec{n_{p_2}}\;\vec{l_{p_2}}))\;-\; log(e^{R_{p_2}}(\lambda))\;\\                  &\;\;\;-\;log(\rho^{R_{p_2}}(\lambda))\;-\;log(m(\vec{n_{p_1}}\;\vec{l_{p_1}}))\;\\
          &\;\;\;-\; log(e^{G_{p_2}}(\lambda))\; -\;log(\rho^{G_{p_1}}(\lambda))\;,\\
\end{aligned}
\end{equation}

\noindent We can factor out the terms $log(e^{R_{p_2}}(\lambda))$ and $log(e^{G_{p_2}}(\lambda))$:

\begin{equation} \label{eq:CCR_step2_suppl}
\begin{aligned}
        log(M_{RG}) &=\;\;\;\;log(m(\vec{n_{p_1}}\;\vec{l_{p_1}}))\;+\;log(\rho^{R_{p_1}}(\lambda))\;\\
        &\;\;\;+\;log(m(\vec{n_{p_2}}\;\vec{l_{p_2}}))\;+\;log(\rho^{G_{p_2}}(\lambda))\;\\
        &\;\;\;-\;log(m(\vec{n_{p_2}}\;\vec{l_{p_2}}))\;-\;log(\rho^{R_{p_2}}(\lambda))\;\\
        &\;\;\;-\;log(m(\vec{n_{p_1}}\;\vec{l_{p_1}}))\;-\;log(\rho^{G_{p_1}}(\lambda))\;,\\
\end{aligned}
\end{equation}

\noindent from Eq~\eqref{eq:CCR_step2_suppl}, the terms $log(m(\vec{n_{p_1}}\;\vec{l_{p_1}}))$ and $log(m(\vec{n_{p_2}}\;\vec{l_{p_2}}))$ are factored out, even though for a curved surface, $\vec{n_{p_1}}\;\neq\;\vec{n_{p_2}}$. Therefore, we have:

\begin{equation} \label{eq:CCR_final_suppl}
\begin{aligned}
        log(M_{RG}) &=\;\;\;\;log(\rho^{R_{p_1}}(\lambda))\;+\;log(\rho^{G_{p_2}}(\lambda)) \\
          &\;\;\;-\;log(\rho^{R_{p_2}}(\lambda))\;-\;log(\rho^{G_{p_1}}(\lambda))\;.\\
\end{aligned}
\end{equation}

\noindent which is the illuminant invariant CCR descriptor for the $R$ \& $G$ channels even with curved surfaces. The other channel $R$ \& $B$ and $G$ \& $B$ pairs can be similarly verified.

\section{Appendix B: Network Configuration Details}
Before we provide the details of the module's technical configuration, we describe the basic building blocks. The network consists of three major structures: an encoder block, an attention layer and a decoder block.

\subsection{Basic Blocks}

\begin{figure}
    \centering
    \includegraphics[width=0.6\linewidth]{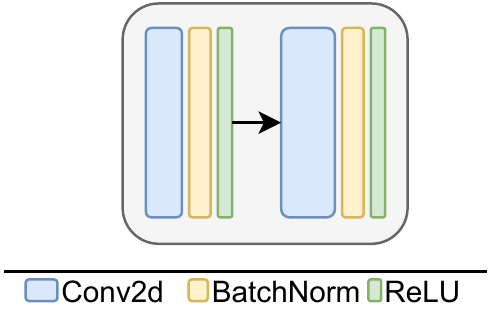}
    \caption{A basic encoder block}
    \label{fig:encoder_block}
\end{figure}

\paragraph{Encoder:}The basic building block of the encoder consists of a 2D convolution layer, a Batch Normalization Layer and a ReLU layer, repeated twice. These 6 layers together create a single block. Fig.~\ref{fig:encoder_block} visualises such a block.

\begin{figure}
    \centering
    \includegraphics[width=0.6\linewidth]{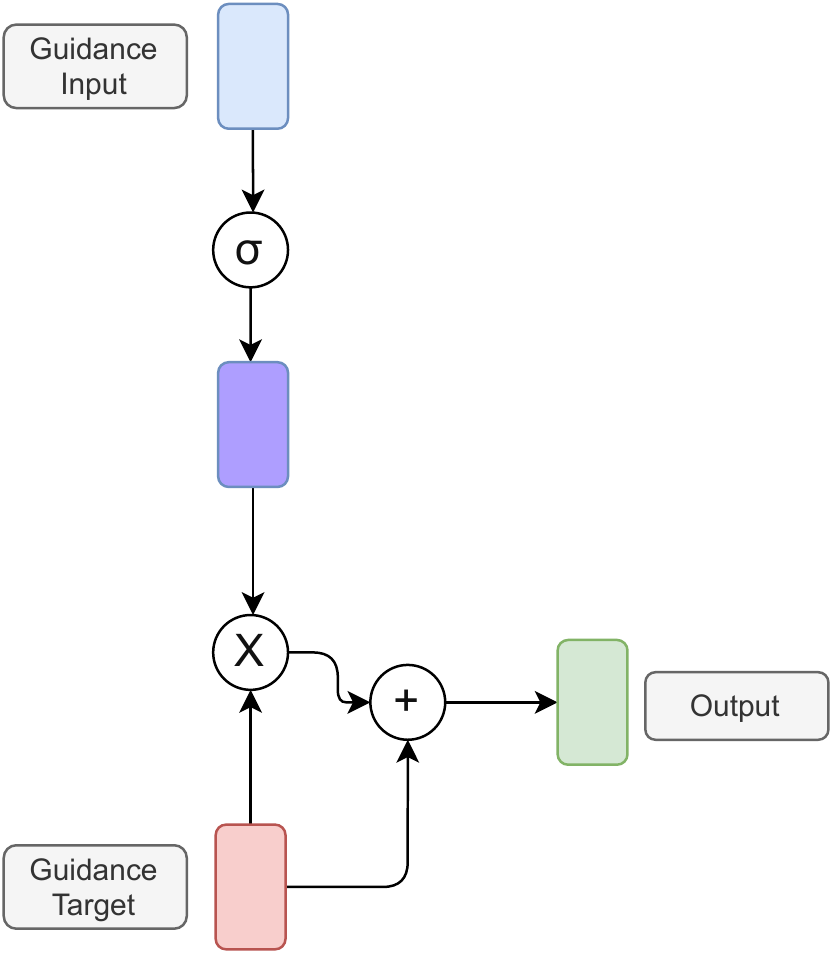}
    \caption{Basic configuration of an attention layer.}
    \label{fig:attention_layer}
\end{figure}

\paragraph{Attention layer:}The attention layer can be defined as:

\begin{equation}
\begin{aligned}
    \mathcal{F}_{int} &= \sigma(\mathcal{F}_{i}) \cdot \mathcal{F}_{t}\;\\
    \mathcal{F}_{attn} &= \mathcal{F}_{int} + \mathcal{F}_{t}
\end{aligned}
\label{eq:attention_layers}
\end{equation}

\noindent where, $\mathcal{F}_{i}$ is the guidance input to the layer, for example, the edge maps. $\mathcal{F}_{t}$ is target values that are to be guided through the attention, for example, the unrefined outputs. $\mathcal{F}_{int}$ is the intermediate output in the attention layer. And $\mathcal{F}_{attn}$ is the final attention guided output. A visual representation can be found in Fig.~\ref{fig:attention_layer}. All the operations are done element-wise.

\begin{figure}
    \centering
    \includegraphics[width=0.6\linewidth]{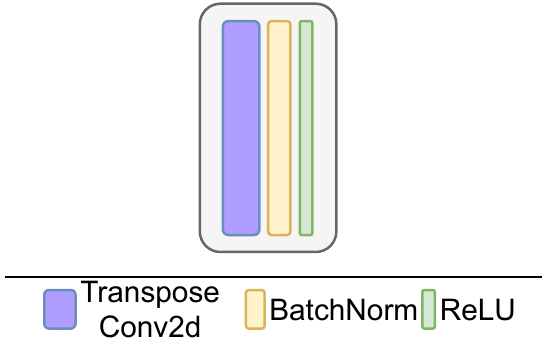}
    \caption{A basic decoder block.}
    \label{fig:decoder_block}
\end{figure}

\paragraph{Decoder:}The basic building block of the decoder consists of a 2D Transposed Convolution layer, a Batch Normalization Layer and a ReLU layer. Fig.~\ref{fig:decoder_block} visualises a single block.

In the following section, we list the details of each of the modules. We list the feed forward configuration in a tabular form. Each of the convolutions and transposed convolutions are always followed by a batch norm and ReLU layer. For brevity, the latter two layers are omitted from the tables.

\subsubsection{Shared Image \& CCR Encoder:}

The configuration details for the image and CCR encoder is provided in table~\ref{tab:img_ccr_encoder}. The same configuration is used for both the image and CCR encoder.

\begin{table}
\centering
\resizebox{0.48\textwidth}{!}{%
\begin{tabular}{|l|l|l|l|}
\hline
Name       & \textbf{Layer} & \textbf{Kernel Size, Stride, Padding} & \textbf{Output Size} \\ \hline
Input      & conv1 & 3x3x64, 1, 1                 & 256x256x64  \\ \hline
           & conv1 & 3x3x64, 1, 1                 & 256x256x64  \\ \hline
           & conv2 & 3x3x64, 2, 1                 & 128x128x64  \\ \hline
           & conv2 & 3x3x128, 1, 1                & 128x128x128 \\ \hline
           & conv3 & 3x3x128, 2, 1                & 64x64x128   \\ \hline
           & conv3 & 3x3x256, 1, 1                & 64x64x256   \\ \hline
           & conv4 & 3x3x256, 2, 1                & 32x32x256   \\ \hline
           & conv4 & 3x3x512, 1, 1                & 32x32x512   \\ \hline
           & conv5 & 3x3x512, 2, 1                & 16x16x512   \\ \hline
Bottleneck & conv5 & 3x3x512, 1, 1                & 16x16x512   \\ \hline
\end{tabular}%
}
\caption{Overview of our encoder configuration, used for both the image and ccr encoders.}
\label{tab:img_ccr_encoder}
\end{table}

\subsubsection{Linked Edge Decoder}

Table~\ref{tab:edge_decoder} lists the configuration for the edge decoder. The $*$ represents the incoming interconnection from the parallel decoder, while $+$ represents the incoming connections through skip connections. All the incoming features are concatenated depth wise before being passed onto the convolution blocks.

\begin{table}
\centering
\resizebox{0.48\textwidth}{!}{%
\begin{tabular}{|l|l|l|l|}
\hline
\textbf{Name} & \textbf{Layer} & \textbf{Kernel Size, Stride, Padding} & \textbf{Output Size} \\ \hline
BottleNeck & deconv1 & 4x4x(512 * 2), 2, 1             & 32x32x512   \\ \hline
           & deconv2 & 4x4x(512 * 2 + 512 + 512), 2, 1 & 64x64x512   \\ \hline
           & deconv3 & 4x4x(512 * 2 + 256 + 256), 2, 1 & 128x128x256 \\ \hline
           & deconv4 & 4x4x(256 * 2 + 128 + 128), 2, 1 & 256x256x128 \\ \hline
           & conv6   & 3x3x(128 * 2 + 64 + 64), 1, 1   & 256x256x64  \\ \hline
Output     & conv6   & 3x3x3, 1, 1                     & 256x256x3   \\ \hline
\end{tabular}%
}
\caption{Overview of the configuration for the edge decoders. The summations represent the skip connections, while the product represents the interconnections between the decoders, i.e., reflectance edge and shading edge decoders.}
\label{tab:edge_decoder}
\end{table}

To add additional feature and scale space supervision to the edges, we also add a multi-scale supervision. For this we transform the intemediate features directly to an output image of size $64$ \& $128$ and add a supervision of the corresponding scaled ground truths. Since we want to have an implicit supervision on the features themselves, we need to minimise the influence of the parameters of this transform. To enforce this, we propose to use the same ``side output" convolution for both the reflectance edge and shading edges. This makes the convolution only learn a common transformation from feature space to image space. The configuration for side outputs are detailed in table~\ref{tab:side_outputs}.

\begin{table}
\centering
\resizebox{0.48\textwidth}{!}{%
\begin{tabular}{|l|l|l|l|}
\hline
Name       & \textbf{Layer} & \textbf{Kernel Size, Stride, Padding} & \textbf{Output Size} \\ \hline
64x64 Output      & conv1 & 3x3x512, 1, 1                 & 64x64x3  \\ \hline
128x128 Output           & conv2 & 3x3x256, 1, 1                 & 128x128x3  \\ \hline
\end{tabular}%
}
\caption{Overview of our side output configuration, used for both the reflectance edge and shading edge decoder features.}
\label{tab:side_outputs}
\end{table}

\subsubsection{Unrefined Decoder}

Table~\ref{tab:unrefined_decoder} shows the configuration for the unrefined decoder. The decoder takes the corresponding edge decoder outputs as the guidance. Skip connections from the image encoder is also provided for colour information. The input to the decoder is the bottleneck from the image encoder. The decoder outputs unrefined reflectance and unrefined shadings, which are globally consistent, but contains local artefacts and physical inconsistencies.

\begin{table}
\centering
\resizebox{0.48\textwidth}{!}{%
\begin{tabular}{|l|l|l|l|}
\hline
\textbf{Name} & \textbf{Layer} & \textbf{Kernel Size, Stride, Padding}                                                            & \textbf{Output Size} \\ \hline
BottleNeck & deconv1   & 4x4x(512 * 1), 2, 1       & 32x32x512   \\ \hline
              & Attention      & \begin{tabular}[c]{@{}l@{}}reflect edge (re) deconv1\\ \& shading edge (se) deconv1\end{tabular} & 32x32x512            \\ \hline
           & deconv2   & 4x4x(512 * 2 + 512), 2, 1 & 64x64x512   \\ \hline
           & Attention & re deconv2 \& se deconv2  & 64x64x512   \\ \hline
           & deconv3   & 4x4x(512 * 2 + 256), 2, 1 & 128x128x256 \\ \hline
           & Attention & re deconv3 \& se deconv3  & 128x128x256 \\ \hline
           & deconv4   & 4x4x(256 * 2 + 128), 2, 1 & 256x256x128 \\ \hline
           & Attention & re deconv4 \& se decovn4  & 256x256x128 \\ \hline
           & conv6     & 3x3x(128 * 2 + 64), 1, 1  & 256x256x64  \\ \hline
           & conv6     & 3x3x3, 1, 1               & 256x256x3   \\ \hline
Output     & Attention & re output \& se output    & 256x256x3   \\ \hline
\end{tabular}%
}
\caption{Overview of the configuration for the unrefined decoders. The summations represent the skip connections, while the product represents the interconnections between the decoders, i.e., unrefined reflectance and shading decoders. The attention layers take two inputs from the edge decoders.}
\label{tab:unrefined_decoder}
\end{table}

\subsubsection{Local Refinement Module}

\begin{figure*}
    \centering
    \includegraphics[width=\linewidth]{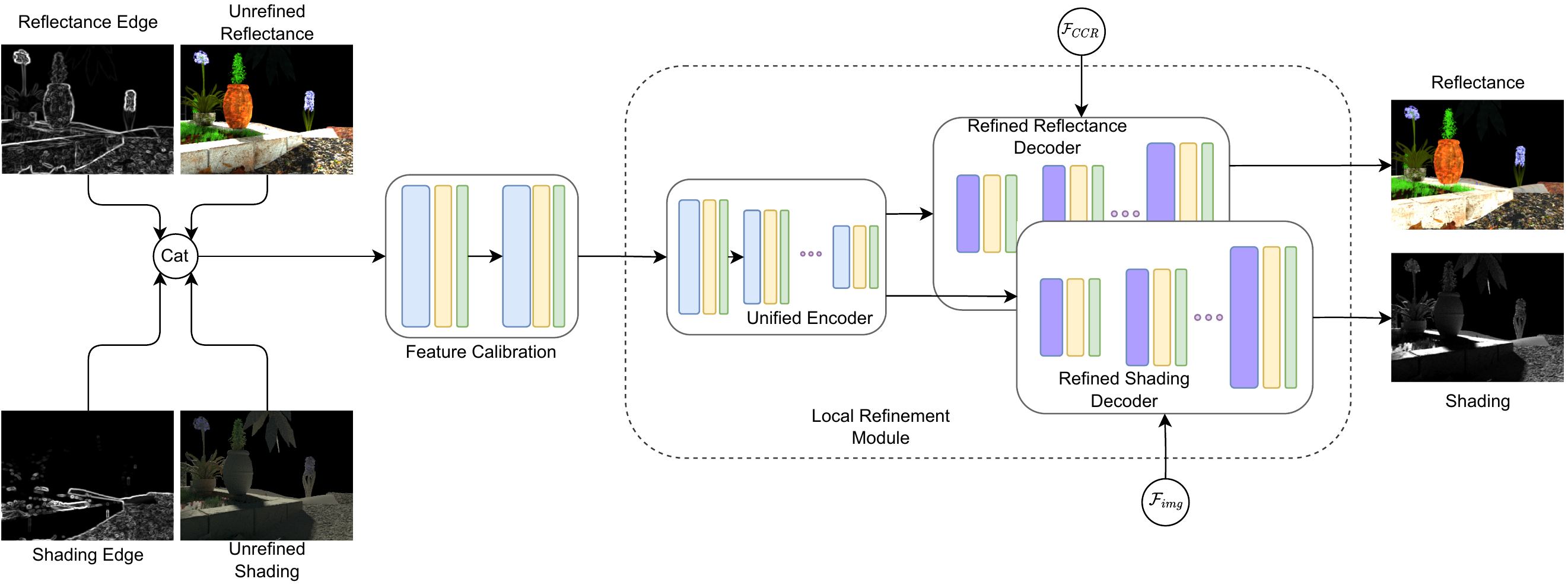}
    \caption{Overview of the local refinement module.}
    \label{fig:refinement_suppl}
\end{figure*}

The configuration for our feature calibration layer is shown in table~\ref{tab:feature_calibrator}. Fig.~\ref{fig:refinement_suppl} shows the refinement module overview. The Reflectance input is the concatenation of the unrefined reflectance (3 channels) and the reflectance edges (3 channels). Similarly, the Shading input is unrefined shading (1 channel) and shading edges (3 channels). We pass it through a $1x1$ convolution to obtain 16 channels each, allowing the network to select and expand the regions that needs most corrections. The transformation does not have any spatial dimension change.

\begin{table}
\centering
\resizebox{0.5\textwidth}{!}{%
\begin{tabular}{|c|c|l|l|}
\hline
Name & \textbf{Layer} & \multicolumn{1}{c|}{\textbf{Kernel Size, Stride, Padding}} & \multicolumn{1}{c|}{\textbf{Output Size}} \\ \hline
\begin{tabular}[c]{@{}c@{}}Reflectance\\ Input\end{tabular}      & reflec conv1 & 1x1x(3 + 3), 1, 0 & 256x256x8  \\ \hline
\begin{tabular}[c]{@{}c@{}}Reflectance\\ Bottleneck\end{tabular} & reflec conv1 & 1x1x16, 1, 0      & 256x256x16 \\ \hline
\begin{tabular}[c]{@{}c@{}}Shading\\ Input\end{tabular}          & shd conv1    & 1x1x(1 + 3), 1, 0 & 256x256x8  \\ \hline
\begin{tabular}[c]{@{}c@{}}Shading\\ Bottleneck\end{tabular}     & shd conv1    & 1x1x16, 1, 0      & 256x256x16 \\ \hline
\end{tabular}%
}
\caption{Overview of the feature calibrator. It has two separate 1x1 convolutions for the reflectance and shading paths. The respective inputs are the concatenated unrefined reflectance \& reflectance edge and unrefined shading \& shading edges.}
\label{tab:feature_calibrator}
\end{table}

The calibrated features are then fed into the refiner encoder to prepare for the refined decoder. The configuration is detailed in table~\ref{tab:unrefined_encoder}.

\begin{table}
\centering
\resizebox{0.48\textwidth}{!}{%
\begin{tabular}{|l|l|l|l|}
\hline
Name       & \textbf{Layer} & \textbf{Kernel Size, Stride, Padding} & \textbf{Output Size} \\ \hline
Input      & conv1 & 3x3x32, 1, 1                 & 256x256x64  \\ \hline
           & conv1 & 3x3x64, 1, 1                 & 256x256x64  \\ \hline
           & conv2 & 3x3x64, 2, 1                 & 128x128x64  \\ \hline
           & conv2 & 3x3x64, 1, 1                & 128x128x128 \\ \hline
           & conv3 & 3x3x128, 2, 1                & 64x64x128   \\ \hline
           & conv3 & 3x3x128, 1, 1                & 64x64x256   \\ \hline
           & conv4 & 3x3x256, 2, 1                & 32x32x256   \\ \hline
           & conv4 & 3x3x256, 1, 1                & 32x32x512   \\ \hline
           & conv5 & 3x3x512, 2, 1                & 16x16x512   \\ \hline
Bottleneck & conv5 & 3x3x512, 1, 1                & 16x16x512   \\ \hline
\end{tabular}%
}
\caption{Overview of the refiner encoder. It takes the calibrated unrefined reflectance and shading as the inputs.}
\label{tab:unrefined_encoder}
\end{table}

The bottleneck from the refiner encoder is then passed onto the local refinement decoder, as shown in table~\ref{tab:refined_decoder}. The output of this decoder is the final IID outputs. The decoder gets skip connections from the refiner encoder. In addition, the skip connections from the first image and CCR encoders are passed through the attention layer before being added to the decoder. Both the image and CCR encoders are separately used as a guidance for the corresponding refiner encoders before being passed to the decoder. This enforces a local level correction, where the feature corresponding CCR and image features are used for the local unrefined outputs.

\begin{table}
\centering
\resizebox{0.48\textwidth}{!}{%
\begin{tabular}{|l|l|l|l|}
\hline
\textbf{Name} & \textbf{Layer} & \textbf{Kernel Size, Stride, Padding}                                    & \textbf{Output Size} \\ \hline
BottleNeck & deconv1   & 4x4x512, 2, 1                   & 32x32x512   \\ \hline
              & Attention      & \begin{tabular}[c]{@{}l@{}}img enc conv4\\ \& unref enc conv4\end{tabular} & 32x32x512            \\ \hline
              & Attention      & \begin{tabular}[c]{@{}l@{}}ccr enc conv4\\ \& unref enc conv4\end{tabular} & 32x32x512            \\ \hline
           & deconv2   & 4x4x(512 * 2 + 512 + 512 + 512), 2, 1 & 64x64x512   \\ \hline
           & Attention & \begin{tabular}[c]{@{}l@{}}img enc conv3\\ \& unref enc conv3\end{tabular}              & 64x64x512   \\ \hline
           & Attention      & \begin{tabular}[c]{@{}l@{}}ccr enc conv4\\ \& unref enc conv4\end{tabular} & 32x32x512            \\ \hline
           & deconv3   & 4x4x(512 * 2 + 256 + 256 + 256), 2, 1 & 128x128x256 \\ \hline
           & Attention & \begin{tabular}[c]{@{}l@{}}img enc conv2\\ \& unref enc conv2\end{tabular}              & 128x128x256 \\ \hline
           & Attention      & \begin{tabular}[c]{@{}l@{}}ccr enc conv4\\ \& unref enc conv4\end{tabular} & 32x32x512            \\ \hline
           & deconv4   & 4x4x(256 * 2 + 128 + 128 + 128), 2, 1 & 256x256x128 \\ \hline
           & Attention & \begin{tabular}[c]{@{}l@{}}img enc conv1\\ \& unref enc conv1\end{tabular}              & 256x256x128 \\ \hline
           & Attention      & \begin{tabular}[c]{@{}l@{}}ccr enc conv4\\ \& unref enc conv4\end{tabular} & 32x32x512            \\ \hline
           & conv6     & 3x3x(128 * 2 + 64 + 64 + 64), 1, 1    & 256x256x64  \\ \hline
Output           & conv6     & 3x3x3, 1, 1                           & 256x256x3   \\ \hline
\end{tabular}%
}
\caption{Overview of the configuration for the local refinement decoders. The summations represent the skip connections (from the unrefined encoder, the image encoder and the ccr encoder), while the product represents the interconnections between the decoders, i.e., the refined reflectance and shading decoders.}
\label{tab:refined_decoder}
\end{table}

\section{Appendix C: Loss Functions}

\begin{figure*}
    \centering
    \includegraphics[width=\linewidth]{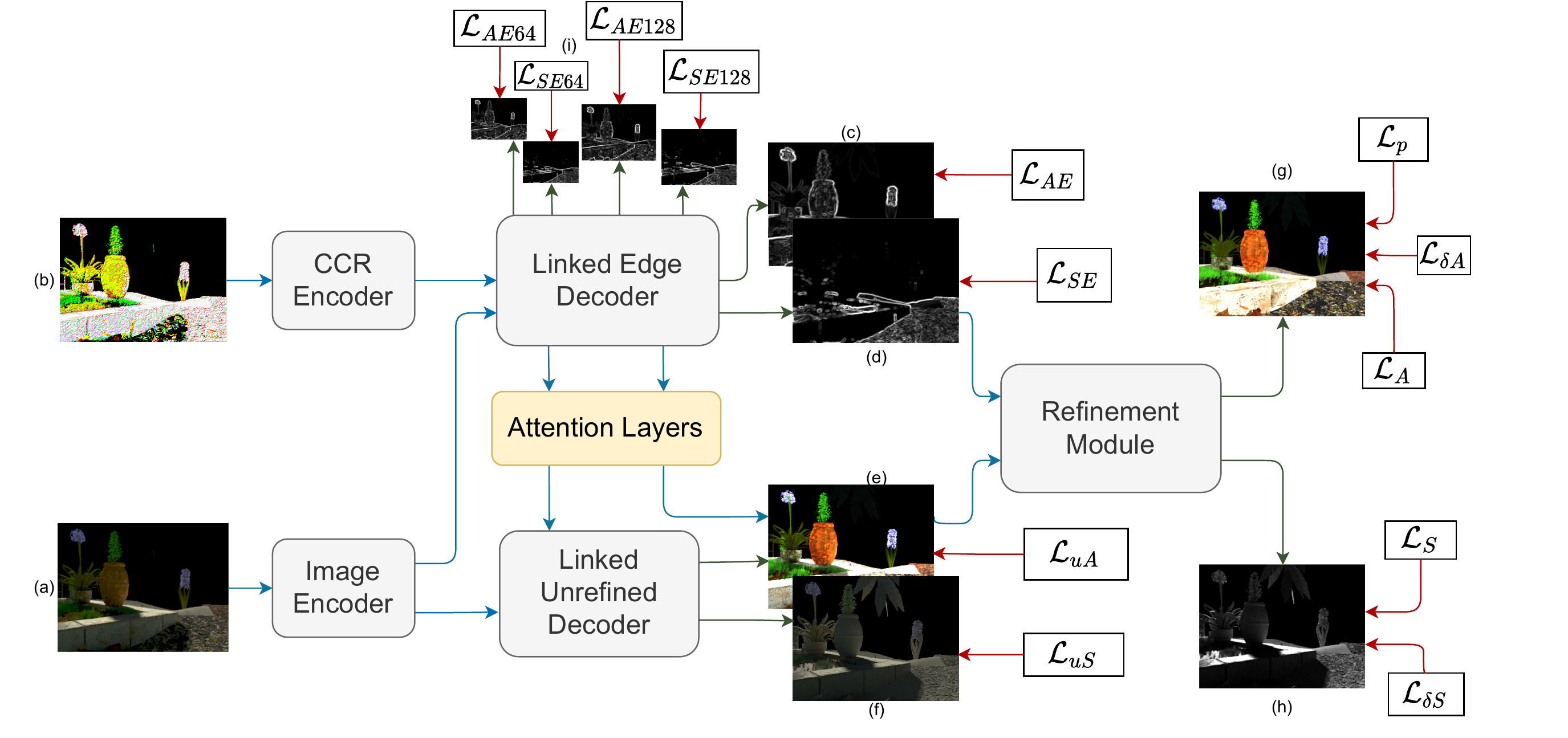}
    \caption{Overview of the supervision enforced on the network. The red arrows show the components on which the corresponding losses are applied. The blue arrows show the dataflow and the green arrows show the outputs of the network.}
    \label{fig:loss_unrefined}
\end{figure*}

To train the network we add explicit supervision to the outputs. For each type of losses, except for the edge, DSSIM and perceptual lossses, we use a combination of scale invariant MSE~\cite{Narihia2015} and the standard MSE loss, as follows:

\begin{equation}
    \mathcal{L}(I, \hat{I}) = \lambda_{smse} \times \mathcal{L}_{SMSE}(I, \hat{I}) + \lambda_{mse} \times \mathcal{L}_{MSE}(I, \hat{I})
    \label{eq:comb_loss}
\end{equation}

\noindent where, $\hat{I}$ is the ground-truth intrinsic image and $I$ is the corresponding predicted image. We empirically set the $\lambda_{SMSE}$ and $\lambda_{MSE}$ to $0.95$ and $0.05$, respectively.

An overview of the losses can be seen in Fig.~\ref{fig:loss_unrefined}. 

The DSSIM is derived from the Structural Similarity Index~\cite{Wang2004} (SSIM) as follows:

\begin{equation}
    \mathcal{L}_{dssim}(I, \hat{I}) = \frac{1 - SSIM(I, \hat{I})}{2}
\end{equation}

\noindent where, $\hat{I}$ is the ground-truth intrinsic image and $I$ is the corresponding predicted image. 

The reconstruction loss is as follows:

\begin{equation}
    \mathcal{L}_{rec} = \mathcal{L} (R\odot S, RGB)
\end{equation}

\noindent where $R$ is the final reflectance output and $S$ is the final shading output of the network. $RGB$ is the input image to the network.

In order to show the influence of the various loss functions, we provide an ablation study on the losses in table~\ref{tab:ablation_loss}.

\begin{table}
\centering
\resizebox{0.48\textwidth}{!}{%
\begin{tabular}{c|c|c|c|c|c|c|}
\cline{2-7}
 &
  \multicolumn{3}{c|}{Reflectance} &
  \multicolumn{3}{c|}{Shading} \\ \cline{2-7} 
 &
  MSE &
  LMSE &
  DSSIM &
  MSE &
  LMSE &
  DSSIM \\ \hline
  \multicolumn{1}{|c|}{\begin{tabular}[c]{@{}c@{}}w/o SSIM\\ Loss\end{tabular}} &
  0.0020 &
  0.0328 &
  0.1766 &
  0.0030 &
  0.0847 &
  0.2314 \\ \hline
  \multicolumn{1}{|c|}{\begin{tabular}[c]{@{}c@{}}w/o Edge\\ Loss\end{tabular}} &
  0.0018 &
  0.0344 &
  0.0846 &
  0.0048 &
  0.1128 &
  0.1973 \\ \hline
  \multicolumn{1}{|c|}{\begin{tabular}[c]{@{}c@{}}w/o Perceptual\\ Loss\end{tabular}} & 0.0019 & 0.0341 & 0.0902 & 0.0022 & 0.0500 & \textbf{0.0791} \\ \hline
  \multicolumn{1}{|c|}{with All losses} &
  \textbf{0.0015} &
  \textbf{0.0289} &
  \textbf{0.0688} &
  \textbf{0.0018} &
  \textbf{0.0489} &
  0.1005 \\ \hline
\end{tabular}%
}
\caption{Ablation study on the various losses. It is shown that removing the losses degrades the performance across the metrics. The drop in performance for the DSSIM metric on the shading component is an acceptable trade-off since that loss improves the convergence of the network from 60 epochs to 15 epochs.}
\label{tab:ablation_loss}
\end{table}

Here we test the performance of the network by removing the DSSIM, the scale space Edge and the Perceptual loss, respectively. We observe that removing the respective losses degrades the performance of the network across all metrics, apart from the Perceptual Loss on the shading DSSIM metric. This is expected since we add the perceptual loss for only the reflectance, since shading for our case is under a white light and hence grayscale. Perceptual loss helps us obtain a better colour as well as structural consistency. Due to the interconnected nature of the components in the network, the perceptual loss would act as a noise source to the shading. Hence, we see a small drop of performance in the structural metric for the shading only, while all the reflectance metrics show an improvement. However, even with the degradation, the performance is still better than the baselines algorithms from the literature, validating the network design choices. Additionally, removing the perceptual loss increases the convergence time from 15 epochs to about 60 epochs. Thus, we reason that this loss of performance on one metric is an acceptable trade-off.

\section{Appendix D: Influence of the type of Attention}

In this section we study the influence of the type of attention. The objective of the network is to learn a transformation of the input image into the intrinsic components. The attention layers are introduced for the network to be able to choose the incoming information. Hence, for our purpose there can be two types of attention: Channel-wise (allowing for cross channel attention, but on a more global contexts) or pixel-wise (spatial attention, allowing for a more finer-grained attention). For the channel wise attention we use the Squeeze and Excitation Network~\cite{Hu2018}. All the hyper-parameters and losses are kept constant as our proposed full pipeline. A numerical comparison is shown in table~\ref{tab:attention_ablation}.

\begin{table}[ht]
\centering

\resizebox{0.48\textwidth}{!}{%
\begin{tabular}{c|c|c|c|c|c|c|}
\cline{2-7}
                                         & \multicolumn{3}{c|}{Reflectance} & \multicolumn{3}{c|}{Shading} \\ \cline{2-7} 
                                         & MSE    & LMSE   & DSSIM  & MSE    & LMSE   & DSSIM  \\ \hline
\multicolumn{1}{|c|}{w/o Attention Layers} & 0.0019    & 0.0330    & 0.0776   & 0.0026   & 0.0704  & 0.1301  \\ \hline
\multicolumn{1}{|c|}{w Channel Attention} & 0.0019    & 0.0335    & 0.0758   & 0.0024   & 0.0638  & 0.1299  \\ \hline
\multicolumn{1}{|c|}{w Spatial Attention}           & \textbf{0.0015} & \textbf{0.0289} & \textbf{0.0688} & \textbf{0.0018} & \textbf{0.0489} & \textbf{0.1005} \\ \hline
\end{tabular}%
}
\caption{We present an ablation study on types of attention layers for our network. From the results, we can see that the spatial attention is better able to model the transformation function. This validates our hypothesis about the need for a finer-grained attention mechanism.}
\label{tab:attention_ablation}
\end{table}

From the table we can see that the channel attention does not offer much improvement. In fact, the very small number of improvements are negligible, compared to without any attention layers. The reflectance component performance is almost like the configuration without any attention layers. Compare this to the spatial attention configuration, which we argue is because there is no single uniform transformation for all the pixels to arrive at the intrinsic components. Having a finer grained attention allows the network to learn a more flexible attentive transformation to arrive at a better decomposition. The numerical results, especially the local metrics (LMSE and DSSIM) that show the most significant improvements, validates this hypothesis.

\section{Appendix E: Note on the performance on the Sintel Dataset}

\begin{figure}
    \centering
    \includegraphics[width=\linewidth]{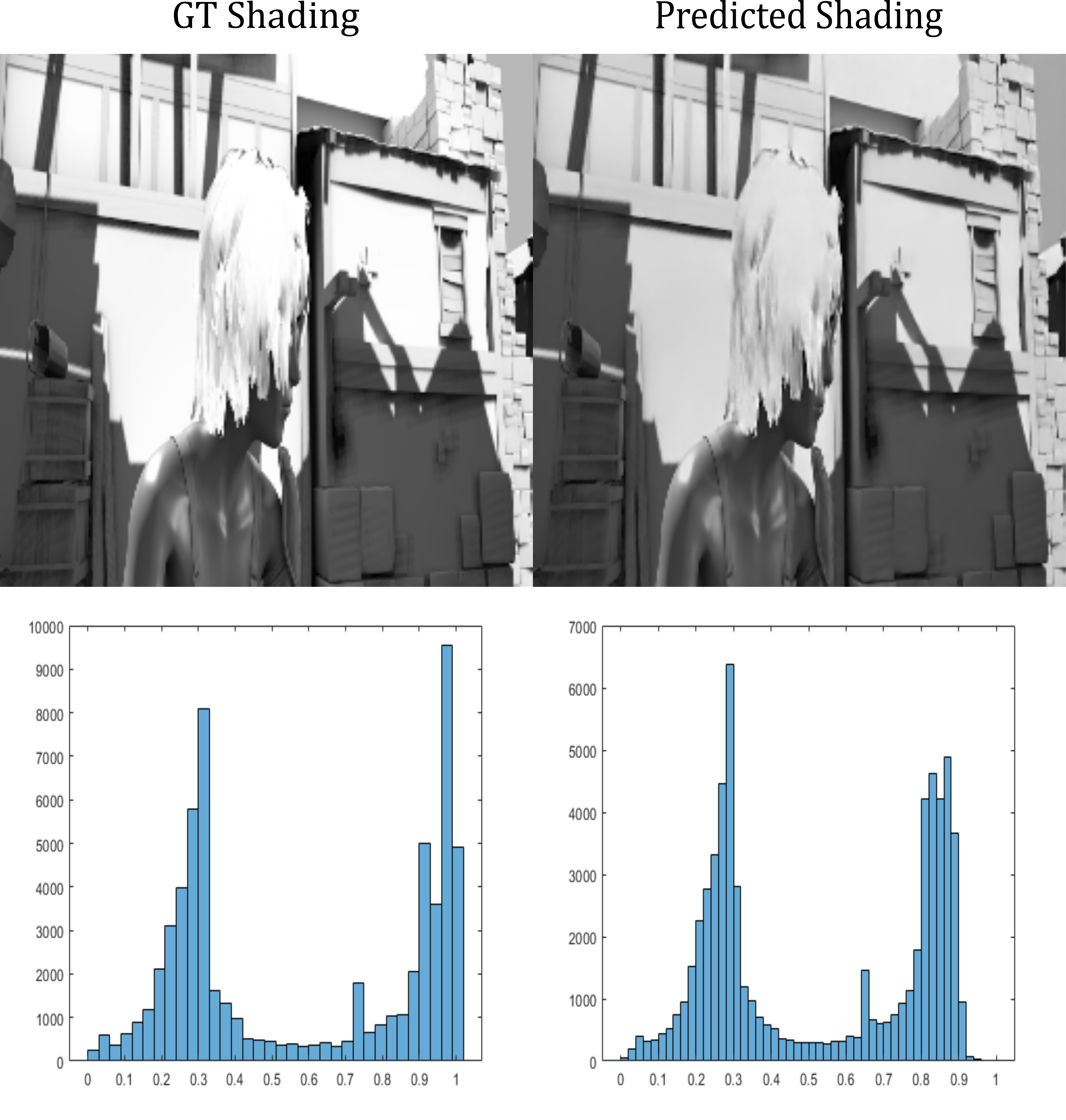}
    \caption{We show the histogram of the predicted shading and the ground truth shading image. It is shown that the predicted shading is of a different scale even though they are structurally similar.}
    \label{fig:sintel_histogram}
\end{figure}

We observe that the network predictions for the shading is a bit darker than the ground truth. However, the outputs look structurally correct. This leads to the lower numbers on the shading MSE and LMSE metrics. Both the metrics measure the euclidean distance between the pixel values and are sensitive to outliers. DSSIM on the other hand considers spatial information and structures, in which we perform better. This is visualised in Fig.~\ref{fig:sintel_histogram}.

From the histogram, we can see that the distribution of the colours is similar, only varying on the scale of the value. From the standard evaluation metrics, the only metric that is sensitive to structures and spatial relationship, is DSSIM. As such, our outputs, even though colour wise and structurally correct, is not an exact match with the GT value pixel wise, resulting in the discrepancy.

Furthermore, our network, by nature is trained on scale invariant, perceptual and SSIM losses. These concentrates more on the perceived colour accuracy and structure than absolute pixel values. For the reflectance, it is not much of a problem since colours are ratios of the RGB channels. Hence, we can see that we are the best performing on the MSE metric and are the second best on the LMSE metric. For the latter metric, this can be explained by the local errors accumulating. However, in case of a grayscale shading (with our white light assumption), the shading is entirely defined as a scaling term. Hence all the scale mismatches amplify the problem, even though structurally the predictions are closer to the ground truth (as shown by the DSSIM metrics).

\section{Appendix F: Extra Visualisations}

\subsection{NED}
\begin{figure*}
    \centering
    \includegraphics[width=0.95\linewidth]{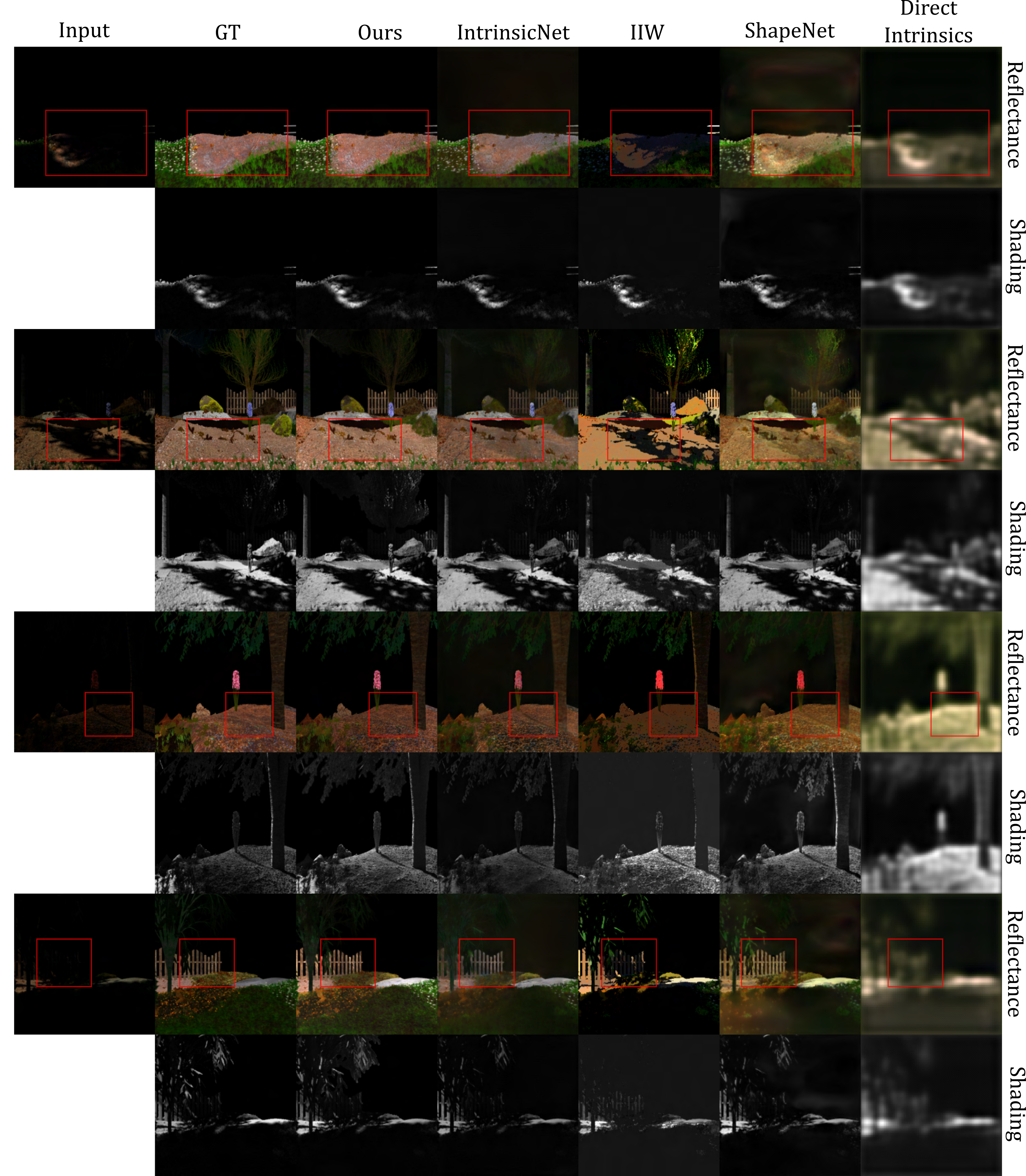}
    \caption{Visuals from the NED test set. It is shown that the network is robust enough to handle cast shadows. In rows 1, 3, 5 and 7, the heavy cast shadows on the ground are completely removed and the predictions are closer to the ground truth images. While IntrinsicNet can also remove the cast shadows, it leaves behind a discolouration in place of the shadows. The other algorithms fail at handling the cast shadows. Input images are gamma corrected for visualisation.}
    \label{fig:ned_1}
\end{figure*}

\begin{figure*}
    \centering
    \includegraphics[width=0.95\linewidth]{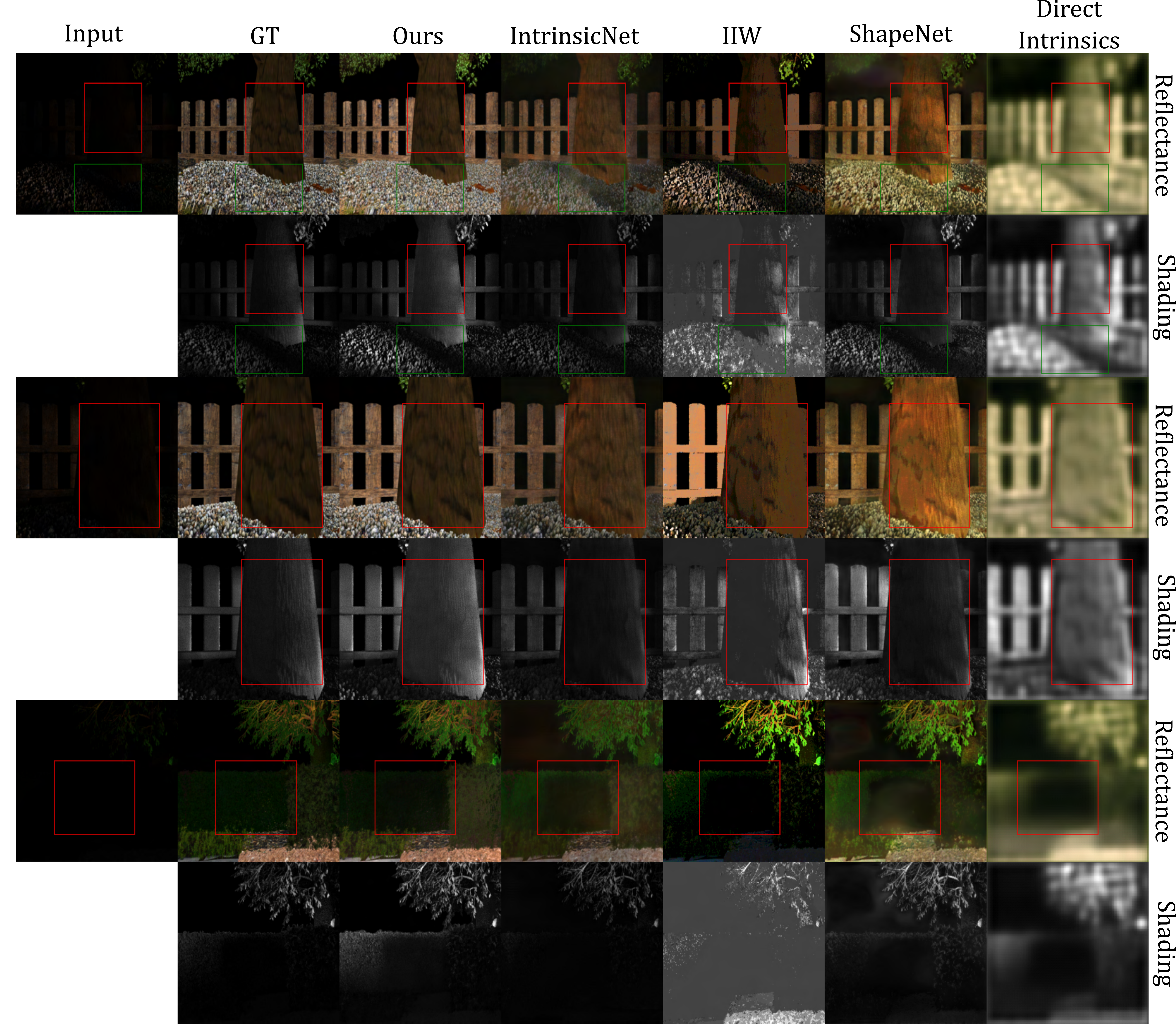}
    \caption{Additional visuals on the NED test set. It is shown that our network, in addition to handling the cast shadow problem, is also capable of preventing texture leakages. Rows 1 and 3 shows a tree trunk that has textures, which is preserved in our prediction, like the ground truth. The other baselines, however, transfer it to the shading images. Additionally, row 5 shows heavy discolouration on the bush, where there was a cast shadow, while our method can recover the reflectance with minimal discolouration. Input images are gamma corrected for visualisation.}
    \label{fig:ned_2}
\end{figure*}

We provide visualisations for the NED test set in Figs.~\ref{fig:ned_1} and~\ref{fig:ned_2}. It is shown that our network can disentangle cast shadows from the reflectance, while also preserving the proper reflectance colour smoothness and textures. The shading is similarly shown to only contain shadows and geometric details only, free from textures.

\subsection{MPI Sintel}
\begin{figure*}
    \centering
    \includegraphics[width=0.8\linewidth]{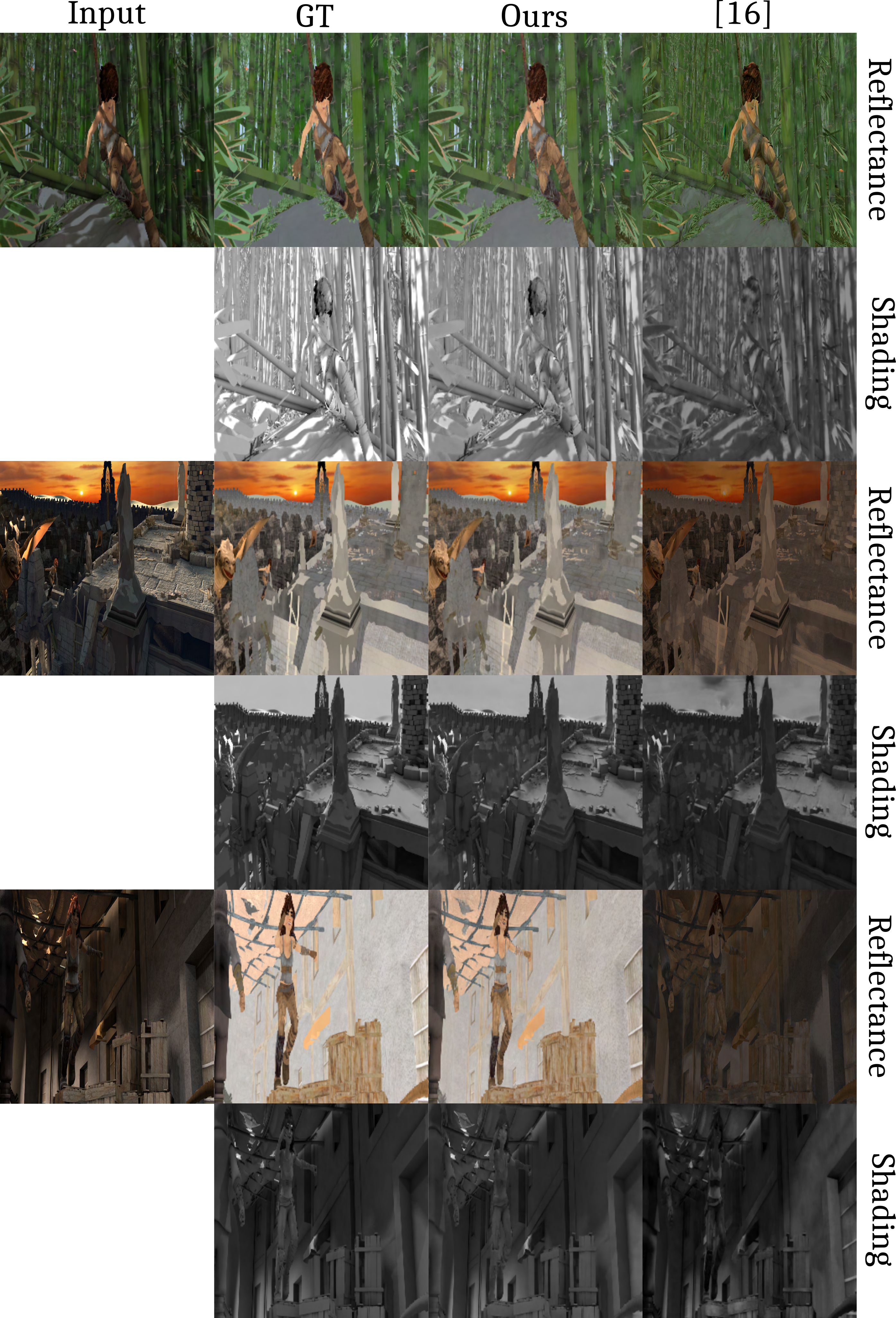}
    \caption{We show visuals on the MPI Sintel test set. It is shown that our network can remove the cast shadows, even from complex scenes like a forest (row 1). On row 5 it is shown that the reflectance is free from the cast shadows on the wall, while the shading image is free from the textures on the wooden box.}
    \label{fig:sintel}
\end{figure*}

\begin{figure*}
    \centering
    \includegraphics[width=0.8\linewidth]{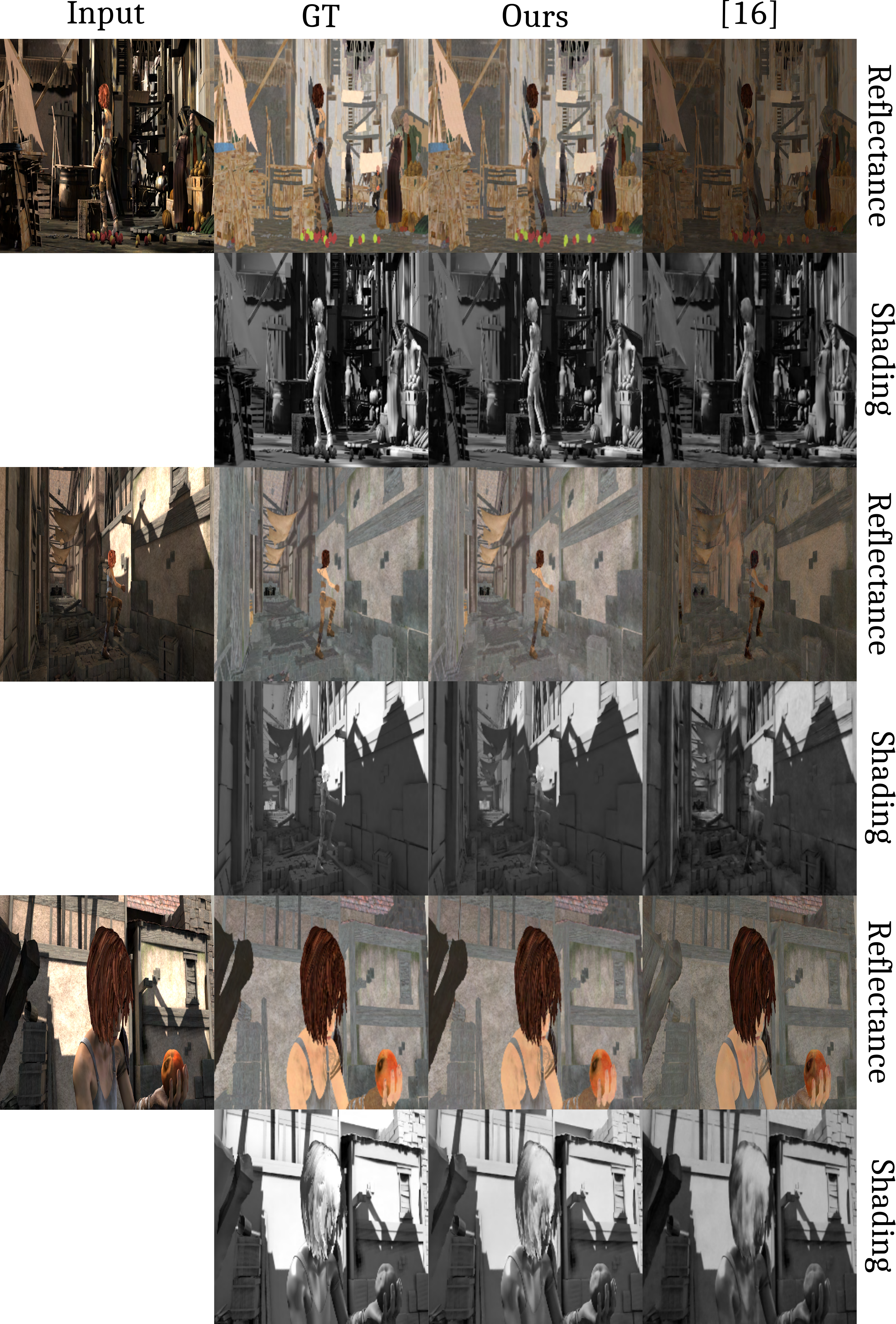}
    \caption{Additional visualisations on the MPI Sintel test set. It is shown that our network is able to handle complex scenes with complicated object interactions and cast shadows.}
    \label{fig:sintel2}
\end{figure*}

We visualise some of the predictions of our network from the test set. The results are visualised in the Figs.~\ref{fig:sintel} and~\ref{fig:sintel2}. It is shown that the predicted outputs are close to the ground truths, robustly handling the cast shadows and textures in the shading and reflectance respectively.

\subsection{MIT Intrinsics}

\begin{figure*}
    \centering
    \includegraphics[width=0.9\linewidth]{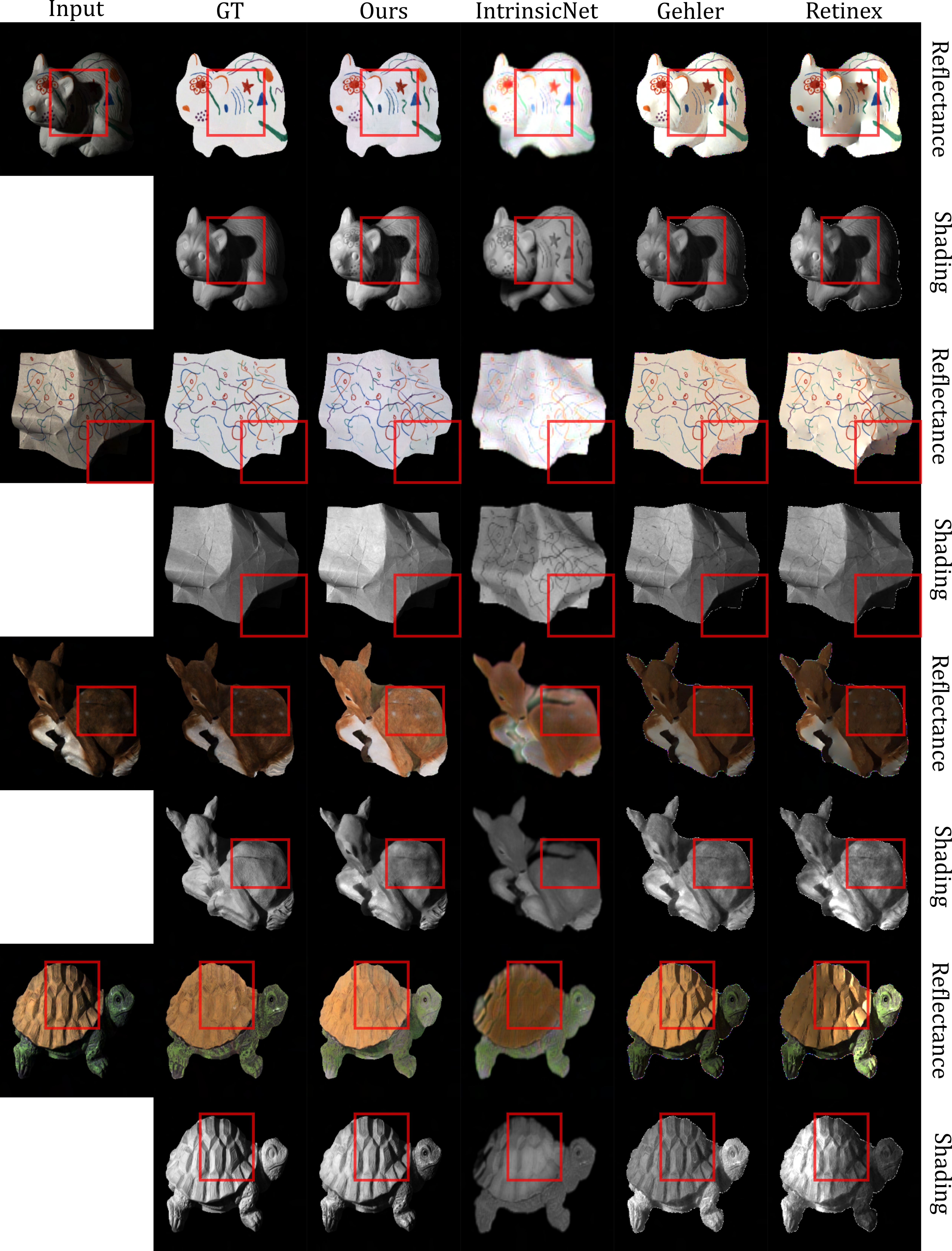}
    \caption{Visuals from the MIT Intrinsic test set. It is shown that the proposed algorithm predictions are closer to the ground truth IID components. IntrinsicNet, on the other hand, completely misses the shadow on the racoon and the paper (rows 2 \& 4), while the proposed algorithm can transfer it to the shading image correctly. The deer and turtle (rows 5 \& 7) show the proposed algorithm able to properly disentangle geometric patterns from reflectance, which are much flatter.}
    \label{fig:mit1}
\end{figure*}

\begin{figure*}
    \centering
    \includegraphics[width=0.8\linewidth]{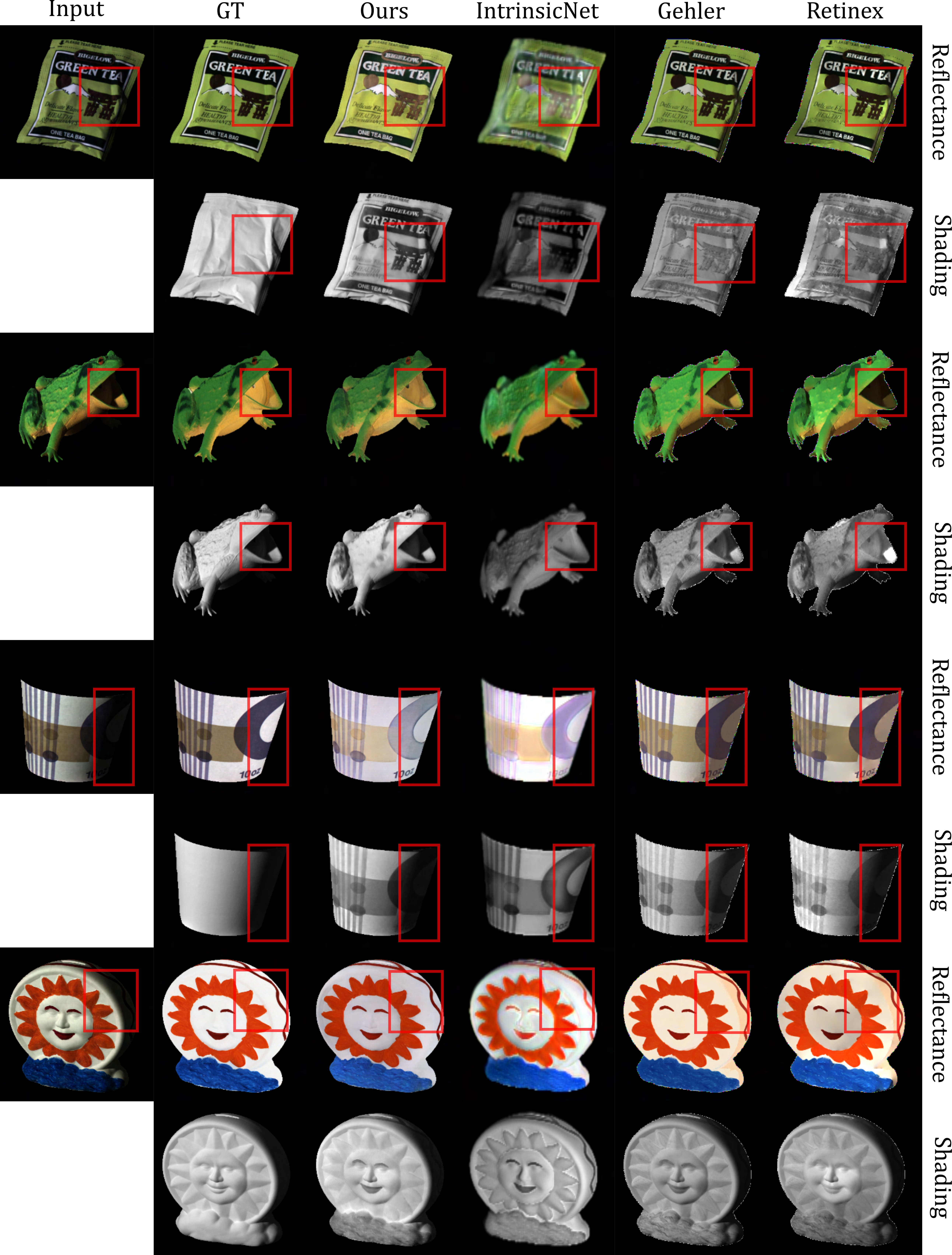}
    \caption{Additional visualisations for the MIT Intrinsic test set. The shadow from the inside of the frog's mouth (row 3) is removed. IntrinsicNet completely misses that (row 4), while other baselines predict it as part of the reflectance. For the cups and sun statue (rows 5 \& 6 and 7 \& 8) the edges with shadow influence (red box) is much flatter for the proposed algorithm, while the baselines have noticeable artefacts due to shadow and reflectance changes.}
    \label{fig:mit2}
\end{figure*}

We present some visualisations of the predictions compared with various baselines on the MIT Intrinsic test set. The visuals are provided in Figs.~\ref{fig:mit1} and~\ref{fig:mit2}. It is shown in the figure that our predictions are much closer to the ground truth compared to the other baselines. IntrinsicNet is shown to have comparatively worse performance, missing shadows (on the raccoon) and often transferring textures in the shadings. In comparison, our method is robust against both leakages, showing the effectiveness of the physics-based guidance.

\subsection{IIW}

\begin{figure*}
    \centering
    \includegraphics[width=0.75\linewidth]{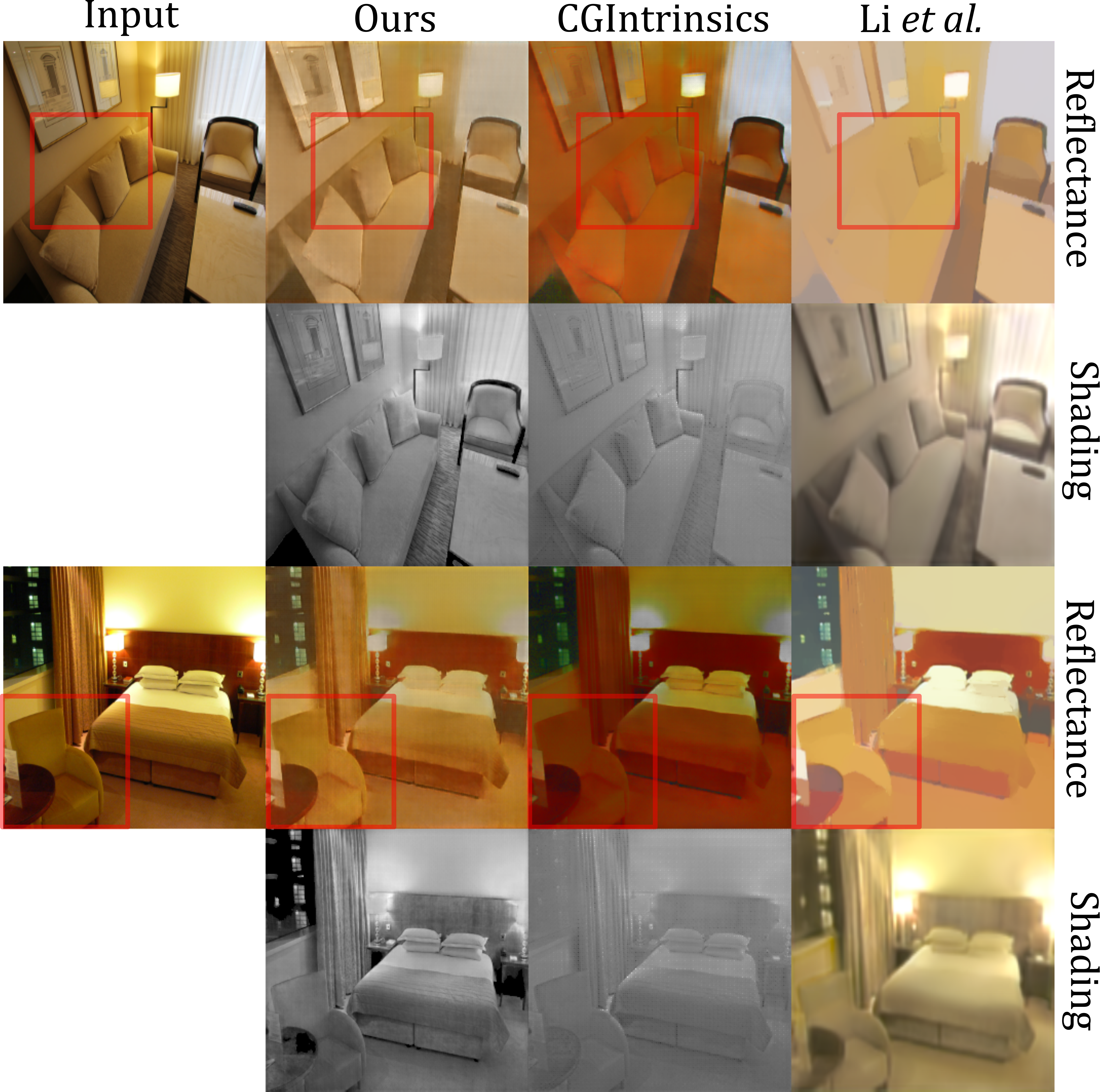}
    \caption{Visuals from the IIW test set. It is shown that our network can predict comparatively better reflectance, which is able to distinguish reflectance boundaries. The proposed network is able to separate the wall from the sofa (first row). Similarly, the competing method classifies the seat of the chair (third row) differently, even though it is part of the same reflectance. The fireplace mantle (fifth row), should be uniform, but the competing methods have uneven reflectance, while our method is able to predict comparatively flatter and uniform reflectance.}
    \label{fig:iiw1}
\end{figure*}

\begin{figure*}
    \centering
    \includegraphics[width=0.7\linewidth]{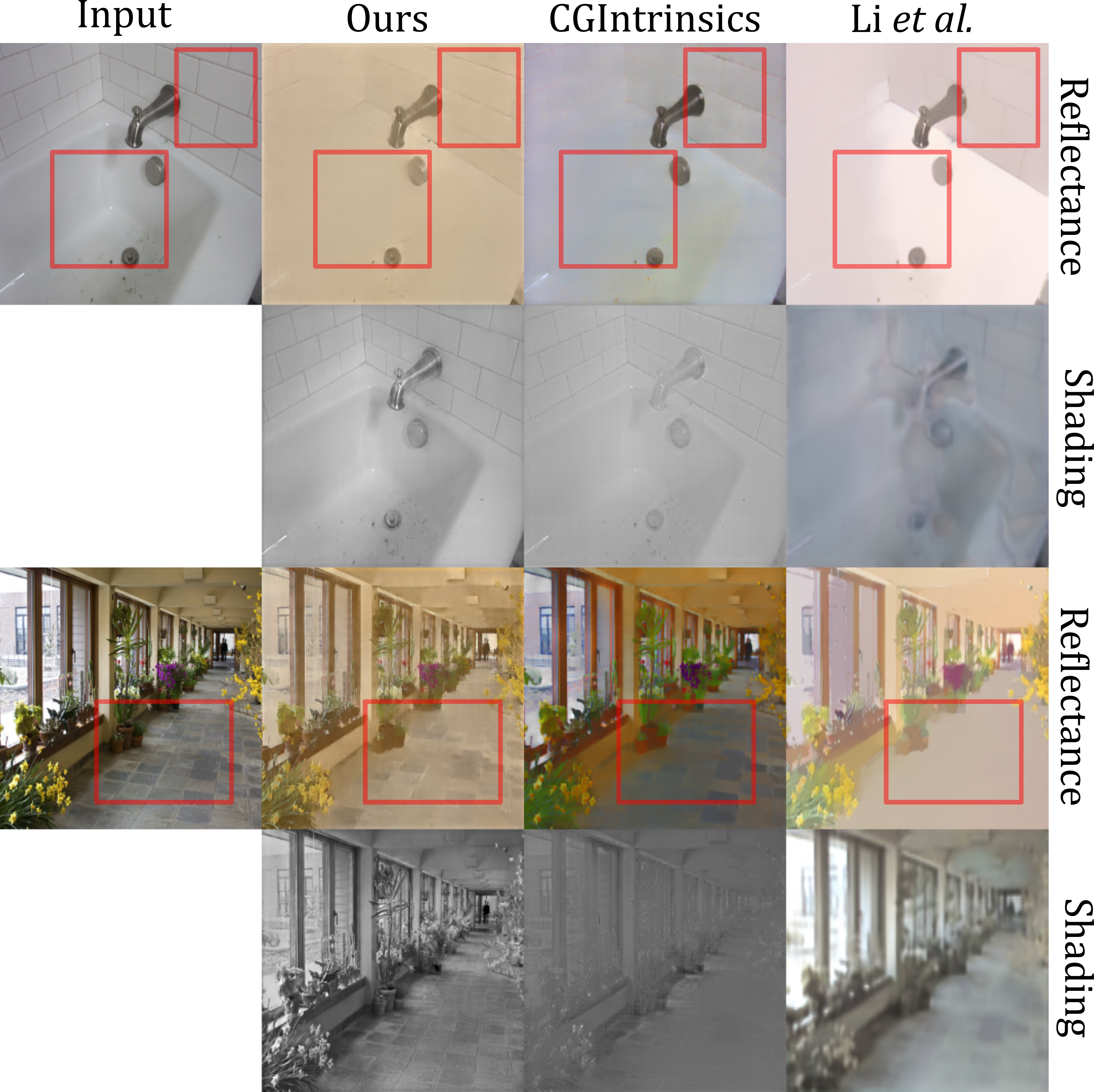}
    \caption{Additional visuals for the IIW test set. It is shown that the predicted network is able to preserve finer details like the individual tiles on the bathroom walls (first row) and floor (third row), while predicting a sharper shading image too. The competing method often predicts a flatter reflectance for all of it, with a blurrier shading image.}
    \label{fig:iiw2}
\end{figure*}

We present visualisations from the IIW test set in Figs.~\ref{fig:iiw1} and~\ref{fig:iiw2}. It is shown that our network can predict reflectances that are consistent with the flatness assumptions. The structural details are correctly transferred to the shadings. CGIntrinsics~\cite{Li2018ECCV} and~\cite{Li2020} on the other hand, misses the reflectance reflectance boundaries. They are also shown to miss finer details like tiles, while also often generating a blurrier shading. The proposed method is able to preserve the tiles and also predict a sharper details, even though it was not trained in the same domain exclusively.

\subsection{Trimbot}

\begin{figure*}
    \centering
    \includegraphics[width=0.55\linewidth]{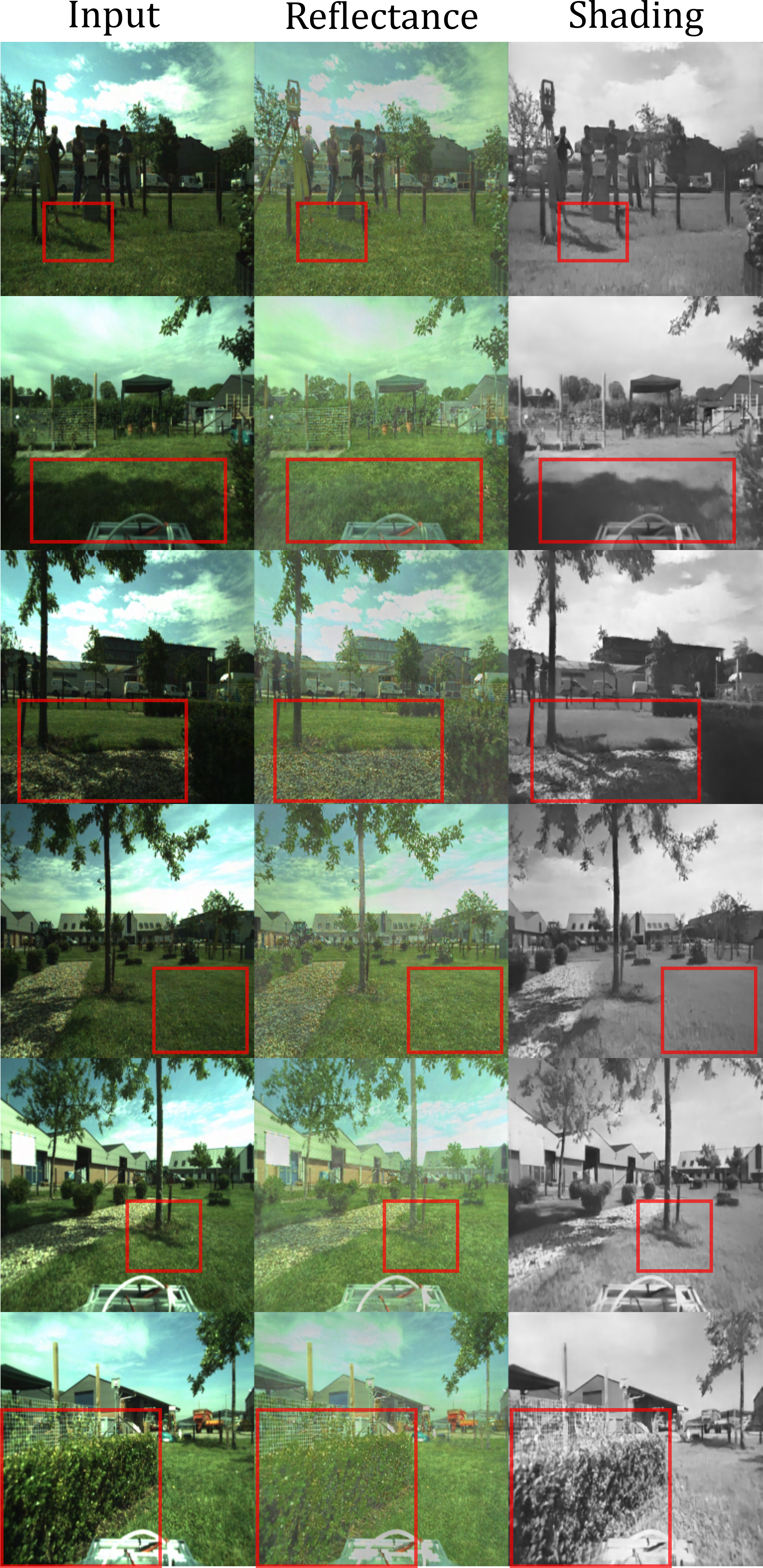}
    \caption{Visuals on the Trimbot dataset. It is shown the proposed algorithm can remove the influence of the cast shadows from the reflectance. In the shading image (row 4), it is shown that the ground is flat and free from textures, due to the flatter geometry of the ground. On the last row, it is shown that the reflectance of the bush also lacks the finer shadows, which show up only in the shading image.}
    \label{fig:trimbot1}
\end{figure*}

To test the effectiveness of our network on real world scenes, we provide outputs on the Trimbot dataset, which are relatively close to our synthetic data settings. The visuals are provided in the Fig.~\ref{fig:trimbot1}. It is shown in the figure that our network can distinguish not only between the shadows and reflectance, but also between the finer objects and shadings. For example, in our outputs, the ground in the shading is flat owing to the flat geometry, while the reflectance shows the reflectance boundary between the grass, thus giving it a rough texture.

\subsection{Real world images}

\begin{figure*}
    \centering
    \includegraphics[width=0.5\linewidth]{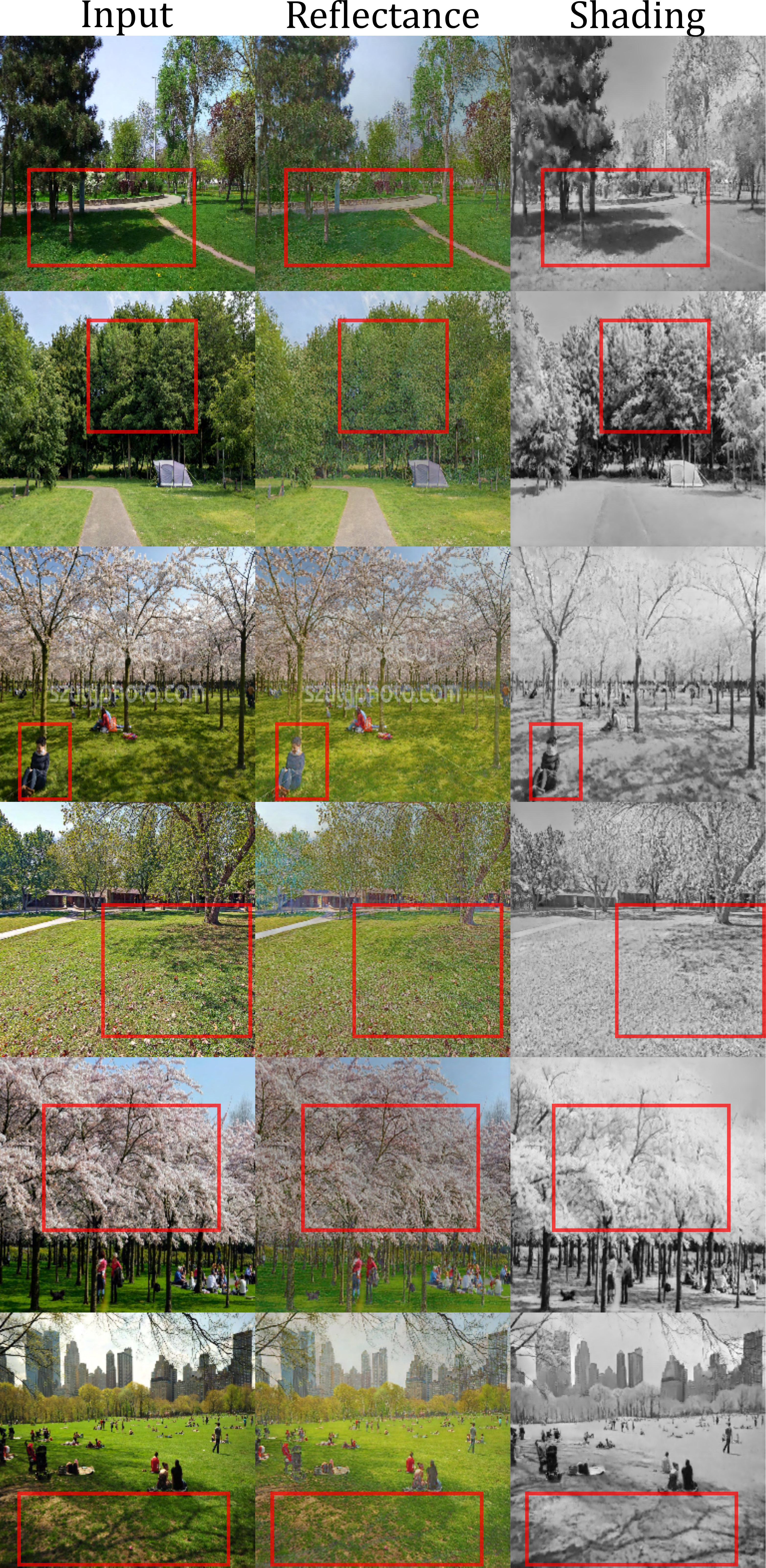}
    \caption{It is shown that the proposed algorithm can model a physics-based formation model, even though it is trained and fine-tuned on purely synthetic data. The shadows on the ground and even among the leaves are removed (rows 1 and 2), while the shading is flat and smooth on the ground, while preserving the small shadows and geometry in the treetops. On the 3rd row, the shadows are mostly removed, while structural details on the woman's shirt is completely flattened in the reflectance.}
    \label{fig:outdoor1}
\end{figure*}

To further test our method, we take a few random internet images. We show the outputs in Fig.~\ref{fig:outdoor1}. It is shown that the network can recover the reflectance, even though it was only trained/fine-tuned on synthetic images. The shadow of the tree on the ground in the first image is separated properly into the reflectance. While in the shading image, the ground is smooth and flat, since the texture is from the reflectance side, while geometrically, the ground is flat and free from variations. Small shadows in the treetops are removed in the reflectance and only preserved in the shading image. Hence, it is shown that the network does not just learn a grayscale transformation for the shading, but also a physics guided IID.

\clearpage

{\small
\bibliographystyle{ieee_fullname}
\bibliography{egbib}

\begin{thebibliography}{10}\itemsep=-1pt

\bibitem{Barron2013}
J.~T. Barron and J. Malik.
\newblock Intrinsic scene properties from a single rgb-d image.
\newblock In {\em CVPR}, 2013.

\bibitem{Barron2015}
J.~T. Barron and J. Malik.
\newblock Shape, illumination, and reflectance from shading.
\newblock {\em IEEE TPAMI}, pages 1670--1687, 2015.

\bibitem{Baslamisli2019}
Anil~S. Baslamisli, Partha Das, Hoang{-}An Le, Sezer Karaoglu, and Theo Gevers.
\newblock Shadingnet: Image intrinsics by fine-grained shading decomposition.
\newblock {\em IJCV}, 129:2445--2473, 2021.

\bibitem{Baslamisli2018ECCV}
A.~S. Baslamisli, T.~T. Groenestege, P. Das, H.~A. Le, S. Karaoglu, and T.
  Gevers.
\newblock Joint learning of intrinsic images and semantic segmentation.
\newblock In {\em ECCV}, 2018.

\bibitem{Baslamisli2018CVPR}
A.~S. Baslamisli, H.~A. Le, and T. Gevers.
\newblock Cnn based learning using reflection and retinex models for intrinsic
  image decomposition.
\newblock In {\em CVPR}, 2018.

\bibitem{Baslamisli2020}
Anil~S. Baslamisli, Yang Liu, Sezer Karaoglu, and Theo Gevers.
\newblock Physics-based shading reconstruction for intrinsic image
  decomposition.
\newblock {\em Comput. Vis. and Image Understanding}, pages 1--14, 2020.

\bibitem{Beigpour2011ObjectRB}
Shida Beigpour and Joost van~de Weijer.
\newblock Object recoloring based on intrinsic image estimation.
\newblock {\em ICCV}, pages 327--334, 2011.

\bibitem{Bell2014}
S. Bell, K. Bala, and N. Snavely.
\newblock Intrinsic images in the wild.
\newblock {\em ACM TOG}, 2014.

\bibitem{Bi2015}
Sai Bi, Xiaoguang Han, and Yizhou Yu.
\newblock An l1 image transform for edge-preserving smoothing and scene-level
  intrinsic decomposition.
\newblock {\em ACM TOG}, 34(4), July 2015.

\bibitem{Bonneel2014}
N. Bonneel, K. Sunkavalli, J. Tompkin, D. Sun, S. Paris, and H. Pfister.
\newblock Interactive intrinsic video editing.
\newblock {\em ACM TOG}, pages 197:1--197:10, 2014.

\bibitem{Bousseau2009}
A. Bousseau, S. Paris, and F. Durand.
\newblock User-assisted intrinsic images.
\newblock {\em ACM TOG}, pages 130:1--130:10, 2009.

\bibitem{Butler2012}
D.~J. Butler, J. Wulff, G.~B. Stanley, and M.~J. Black.
\newblock A naturalistic open source movie for optical flow evaluation.
\newblock In {\em ECCV}, 2012.

\bibitem{Chen2013}
Q. Chen and V. Koltun.
\newblock A simple model for intrinsic image decomposition with depth cues.
\newblock In {\em ICCV}, 2013.

\bibitem{Cheng2018}
L. Cheng, C. Zhang, and Z. Liao.
\newblock Intrinsic image transformation via scale space decomposition.
\newblock In {\em CVPR}, 2018.

\bibitem{Cheng2019}
Ziang Cheng, Yinqiang Zheng, Shaodi You, and Imari Sato.
\newblock Non-local intrinsic decomposition with near-infrared priors.
\newblock In {\em ICCV}, October 2019.

\bibitem{Fan2018}
Q. Fan, J. Yang, G. Hua, B. Chen, and D. Wipf.
\newblock Revisiting deep intrinsic image decompositions.
\newblock In {\em CVPR}, 2018.

\bibitem{Garces2012}
Elena Garces, Adolfo Munoz, Jorge Lopez-Moreno, and Diego Gutierrez.
\newblock Intrinsic images by clustering.
\newblock {\em Comput. Graph. Forum (Proceedings of the Eurographics Symposium
  on Rendering)}, 31(4), 2012.

\bibitem{Gehler2011}
P.~V. Gehler, C. Rother, M. Kiefel, L. Zhang, and B. Schölkopf.
\newblock Recovering intrinsic images with a global sparsity prior on
  reflectance.
\newblock In {\em NeurIPS}, 2011.

\bibitem{Gevers1999}
T. Gevers and A. Smeulders.
\newblock Color-based object recognition.
\newblock {\em PR}, pages 453--464, 1999.

\bibitem{Grosse2009}
R. Grosse, M.~K. Johnson, E.~H. Adelson, and W.~T. Freeman.
\newblock Ground truth dataset and baseline evaluations for intrinsic image
  algorithms.
\newblock In {\em ICCV}, 2009.

\bibitem{henderson2019}
Paul Henderson and Vittorio Ferrari.
\newblock Learning single-image 3d reconstruction by generative modelling of
  shape, pose and shading.
\newblock {\em International Journal of Computer Vision}, 2019.

\bibitem{Hu2018}
J. Hu, L. Shen, and G. Sun.
\newblock Squeeze-and-excitation networks.
\newblock In {\em CVPR}, 2018.

\bibitem{Janner2017}
M. Janner, J. Wu, T.~D. Kulkarni, I. Yildirim, and J.~B. Tenenbaum.
\newblock Self-supervised intrinsic image decomposition.
\newblock In {\em NeurIPS}, 2017.

\bibitem{Jeon2014}
J. Jeon, S. Cho, X. Tong, and S. Lee.
\newblock Intrinsic image decomposition using structure-texture separation and
  surface normals.
\newblock In {\em ECCV}, 2014.

\bibitem{Land1971}
E.~H. Land and J.~J. McCann.
\newblock Lightness and retinex theory.
\newblock {\em J. of Optical Society of America}, pages 1--11, 1971.

\bibitem{Lee2012}
Kyong~Joon Lee, Qi Zhao, Xin Tong, Minmin Gong, Shahram Izadi, Sang~Uk Lee,
  Ping Tan, and Stephen Lin.
\newblock Estimation of intrinsic image sequences from image+depth video.
\newblock In Andrew Fitzgibbon, Svetlana Lazebnik, Pietro Perona, Yoichi Sato,
  and Cordelia Schmid, editors, {\em ECCV}, pages 327--340, Berlin, Heidelberg,
  2012. Springer Berlin Heidelberg.

\bibitem{Lettry2018}
L. Lettry, K. Vanhoey, and L. van Gool.
\newblock Unsupervised deep single-image intrinsic decomposition using
  illumination-varying image sequences.
\newblock In {\em Int. Pacific Conf. on Comput. Graph. and App.}, 2018.

\bibitem{Li2020}
Zhengqin Li, Mohammad Shafiei, R. Ramamoorthi, Kalyan Sunkavalli, and Manmohan
  Chandraker.
\newblock Inverse rendering for complex indoor scenes: Shape, spatially-varying
  lighting and svbrdf from a single image.
\newblock {\em CVPR}, pages 2472--2481, 2020.

\bibitem{Li2018ECCV}
Z. Li and N. Snavely.
\newblock Cgintrinsics: Better intrinsic image decomposition through
  physically-based rendering.
\newblock In {\em ECCV}, 2018.

\bibitem{Li2018CVPR}
Z. Li and N. Snavely.
\newblock Learning intrinsic image decomposition from watching the world.
\newblock In {\em CVPR}, 2018.

\bibitem{Liu2020}
Y. Liu, Y. Li, Shaodi You, and Feng Lu.
\newblock Unsupervised learning for intrinsic image decomposition from a single
  image.
\newblock {\em CVPR}, pages 3245--3254, 2020.

\bibitem{Ma2020}
Y. {Ma}, X. {Jiang}, Z. {Xia}, M. {Gabbouj}, and X. {Feng}.
\newblock Casqnet: Intrinsic image decomposition based on cascaded quotient
  network.
\newblock {\em IEEE TCSVT}, pages 1--1, 2020.

\bibitem{Matsushita2004}
Y. Matsushita, K. Nishino, K. Ikeuchi, and M. Sakauchi.
\newblock Illumination normalization with time-dependent intrinsic images for
  video surveillance.
\newblock {\em IEEE TPAMI}, pages 1336--1347, 2004.

\bibitem{Meka2016}
A. Meka, M. Zollhöfer, C. Richardt, and C. Theobalt.
\newblock Live intrinsic video.
\newblock {\em ACM TOG}, 2016.

\bibitem{Narihia2015}
T. Narihira, M. Maire, and S.~X. Yu.
\newblock Direct intrinsics: Learning albedo-shading decomposition by
  convolutional regression.
\newblock In {\em ICCV}, 2015.

\bibitem{Narihira2015-2}
T. {Narihira}, M. {Maire}, and S.~X. {Yu}.
\newblock Learning lightness from human judgement on relative reflectance.
\newblock In {\em CVPR}, pages 2965--2973, June 2015.

\bibitem{Nestmeyer2016}
Thomas Nestmeyer and Peter~V. Gehler.
\newblock Reflectance adaptive filtering improves intrinsic image estimation.
\newblock {\em CoRR}, abs/1612.05062, 2016.

\bibitem{Sattler2017}
T. Sattler, R. Tylecek, T. Brox, M. Pollefeys, and R.~B. Fisher.
\newblock 3d reconstruction meets semantics - reconstruction challange 2017.
\newblock In {\em Int. Conf. Comput. Vis. Workshop}, 2017.

\bibitem{Sengupta2019}
Soumyadip Sengupta, Jinwei Gu, Kihwan Kim, Guilin Liu, David~W. Jacobs, and Jan
  Kautz.
\newblock Neural inverse rendering of an indoor scene from a single image.
\newblock {\em CoRR}, abs/1901.02453, 2019.

\bibitem{Shafer1985}
S. Shafer.
\newblock Using color to separate reflection components.
\newblock {\em Color Research and App.}, pages 210--218, 1985.

\bibitem{Shen2008}
L. Shen, P. Tan, and S. Lin.
\newblock Intrinsic image decomposition with non-local texture cues.
\newblock In {\em CVPR}, 2008.

\bibitem{Shen2011}
L. Shen and C. Yeo.
\newblock Intrinsic images decomposition using a local and global sparse
  representation of reflectance.
\newblock In {\em CVPR}, 2011.

\bibitem{Shi2017}
J. Shi, Y. Dong, H. Su, and S.~X. Yu.
\newblock Learning non-lambertian object intrinsics across shapenet categories.
\newblock In {\em CVPR}, 2017.

\bibitem{shu2017}
Zhixin Shu, Ersin Yumer, Sunil Hadap, Kalyan Sunkavalli, Eli Shechtman, and
  Dimitris Samaras.
\newblock Neural face editing with intrinsic image disentangling.
\newblock {\em CoRR}, abs/1704.04131, 2017.

\bibitem{Simonyan2015}
K. Simonyan and A. Zisserman.
\newblock Very deep convolutional networks for large-scale image recognition.
\newblock In {\em ICLR}, 2015.

\bibitem{Wada1995}
T. Wada, H. Ukida, and T. Matsuyama.
\newblock Shape from shading with interreflections under proximal light
  source-3d shape reconstruction of unfolded book surface from a scanner image.
\newblock In {\em ICCV}, 1995.

\bibitem{Wang2004}
Z. Wang, A.~C. Bovik, H.~R. Sheikh, and E.~P. Simoncelli.
\newblock Image quality assessment: From error visibility to structural
  similarity.
\newblock {\em IEEE TIP}, pages 600--612, 2004.

\bibitem{Weiss2001}
Y. Weiss.
\newblock Deriving intrinsic images from image sequences.
\newblock In {\em ICCV}, 2001.

\bibitem{Xie2015}
Saining "Xie and Zhuowen" Tu.
\newblock Holistically-nested edge detection.
\newblock In {\em ICCV}, 2015.

\bibitem{Xu2019}
Chen Xu, Yu Han, George Baciu, and Min Li.
\newblock Fabric image recolorization based on intrinsic image decomposition.
\newblock {\em Textile Research J.}, 89(17):3617--3631, 2019.

\bibitem{Xu2020}
J. Xu, Y. Hou, D. Ren, L. Liu, F. Zhu, M. Yu, H. Wang, and L. Shao.
\newblock Star: A structure and texture aware retinex model.
\newblock {\em IEEE TIP}, pages 5022--5037, 2020.

\bibitem{Ye2014}
G. Ye, E. Garces, Y. Liu, Q. Dai, and D. Gutierrez.
\newblock Intrinsic video and applications.
\newblock {\em ACM TOG}, 2014.

\bibitem{Yu2019}
Y. Yu and W.~A.~P. Smith.
\newblock Inverserendernet: Learning single image inverse rendering.
\newblock In {\em CVPR}, 2019.

\bibitem{Yuan2019}
Y. Yuan, B. Sheng, P. Li, L. Bi, J. Kim, and E. Wu.
\newblock Deep intrinsic image decomposition using joint parallel learning.
\newblock In {\em Comput. Graph. Int. Conf.}, 2019.

\bibitem{Zhao2012}
Q. {Zhao}, P. {Tan}, Q. {Dai}, L. {Shen}, E. {Wu}, and S. {Lin}.
\newblock A closed-form solution to retinex with nonlocal texture constraints.
\newblock {\em IEEE TPAMI}, 34(7):1437--1444, July 2012.

\bibitem{Zhou2019}
Hao Zhou, Xiang Yu, and David~W. Jacobs.
\newblock Glosh: Global-local spherical harmonics for intrinsic image
  decomposition.
\newblock In {\em ICCV}, October 2019.

\bibitem{Zhou2015}
Tinghui Zhou, Philipp Kr{\"{a}}henb{\"{u}}hl, and Alexei~A. Efros.
\newblock Learning data-driven reflectance priors for intrinsic image
  decomposition.
\newblock {\em CoRR}, abs/1510.02413, 2015.

\end{thebibliography}
}

\end{document}